\documentclass[sigconf]{acmart}
\AtBeginDocument{%
  }

\setcopyright{none}
\renewcommand\footnotetextcopyrightpermission[1]{}

\usepackage{amsmath}
\usepackage{algorithm}
\usepackage{algpseudocode}
\usepackage{booktabs}
\usepackage{multirow}
\usepackage{makecell}
\usepackage{threeparttable}
\usepackage{tabularx}
\usepackage{xcolor}

\newcommand{\method}{\textsc{XAI-Refine}}

\begin{document}

\title{\method{}: An Automated Explanation--Knowledge Loop for Brain-Age Prediction}

\author{Yang Qiao}
\affiliation{%
  \department{Department of Computer Science}
  \institution{Emory University}
  \city{Atlanta}
  \state{Georgia}
  \country{USA}
}
\email{yqiao47@emory.edu}

\author{Junjie Wu}
\affiliation{%
  \department{Department of Radiology and Imaging Sciences}
  \institution{Computational NeuroImaging \& Neuroscience Lab, Emory University}
  \city{Atlanta}
  \state{Georgia}
  \country{USA}
}
\email{junjie.wu@emory.edu}

\author{Deqiang Qiu}
\affiliation{%
  \department{Department of Radiology and Imaging Sciences}
  \institution{Computational NeuroImaging \& Neuroscience Lab, Emory University}
  \city{Atlanta}
  \state{Georgia}
  \country{USA}
}
\email{deqiang.qiu@emory.edu}

\author{James J. Lah}
\affiliation{%
  \department{Department of Neurology}
  \institution{Emory University School of Medicine}
  \city{Atlanta}
  \state{Georgia}
  \country{USA}
}
\email{jlah@emory.edu}

\author{Liang Zhao}
\affiliation{%
  \department{Department of Computer Science}
  \institution{Emory University}
  \city{Atlanta}
  \state{Georgia}
  \country{USA}
}
\email{liang.zhao@emory.edu}
\renewcommand{\shortauthors}{Qiao et al.}

\begin{abstract}
Brain-age prediction models are commonly evaluated by predictive accuracy,
yet accurate predictions alone do not establish that a model relies on
reproducible or neurobiologically supported mechanisms. Post-hoc explanation
methods can expose these mechanisms, but existing workflows typically stop at
diagnosis or require correction targets to be specified before model analysis.
We propose \method{}, an automated explanation--knowledge loop for brain-age
prediction from resting-state functional connectivity. At each iteration,
\method{} consolidates complementary post-hoc analyses across repeated
training runs into reliable, structured model explanations. It converts each
reliable explanation into a neutral neurobiological question, retrieves and
verifies relevant literature, and compiles the verified evidence into an
admissible set in the same typed explanation space. The target for refinement is defined as the minimal projection of the current
model explanation onto the admissible set induced by applicable verified
knowledge. This revised explanation is then
translated into a differentiable constraint while preserving the originating
model variable, measurement operator, and applicable scope. Candidate updates are promoted only when multi-seed validation confirms
target-directed explanatory movement, predictive performance remains within a
prespecified guardrail, and non-target explanatory drift remains bounded.
Experiments on FC-based brain-age prediction evaluate predictive performance,
explanation reliability, literature alignment, and target-specific model
revision, illustrating a structured route from post-hoc analysis to
evidence-guided model refinement.
\end{abstract}




\maketitle

\section{Introduction}
\label{sec:introduction}

Brain-age prediction estimates chronological age from neuroimaging and is
widely used to characterize age-related variation in the brain. Resting-state
functional connectivity (FC) is attractive for this purpose because it
represents distributed interactions among brain regions and can be modeled as
a structured network. Existing FC-based approaches range from convolutional
predictors to latent-factor and graph models
\cite{li2018brain,monti2019interpretable,gao2023brain}.
However, low prediction error does not establish that a model relies on a
reproducible or neurobiologically supported mechanism. This limitation is
especially important in brain-age studies, where prediction slope, age-related
bias, and longitudinal consistency may differ even between models with similar
MAE \cite{de2021mind}.

Post-hoc explanations expose model behavior, but a single explanation can be
unstable across training runs or weakly coupled to the underlying predictor.
Sanity and benchmarking studies have shown that some attribution procedures
may be insensitive to model parameters or fail to recover important features
reliably \cite{adebayo2018sanity,hooker2019benchmark}. Aggregating repeated or
complementary explanations can improve robustness
\cite{rieger2019aggregating}, yet reproducibility alone is insufficient: a
stable explanation may still describe a stable shortcut rather than an
applicable scientific relation. A scientific workflow must therefore connect
model faithfulness, empirical reproducibility, and independently verified
domain evidence.

Explanation-Guided Learning (EGL) moves beyond diagnosis by incorporating
explanation targets into training \cite{gao2024going}. Representative methods
penalize incorrect input gradients, align attributions with expert priors, or
encode causal and symbolic constraints
\cite{ross2017right,schramowski2020making,
weinberger2019learning,erion2021improving,xu2018semantic,
kancheti2021matching}. These approaches normally assume that the desired
prior or correction target is known before model analysis. Recent LLM-based
systems can retrieve scientific evidence, generate research ideas, or provide
feature priors \cite{wadden2020fact,wadden2022scifact,
baek2024researchagent,vukadin2025large}, but they do not determine how
a verified claim should minimally revise a specific explanation extracted
from a learned scientific predictor.

We propose \method{}, an automated explanation--knowledge loop that derives
rather than prespecifies the refinement target. At each iteration, the current
predictor is analyzed across multiple training runs and explanation families.
Reliable findings are represented as typed records that preserve the model
variable, measurement operator, and applicable scope. A literature-verification
agent converts each record into a neutral scientific question and compiles
applicable evidence into an admissible set in the same explanation space. The
revised explanation is the minimum change to the current explanation that is
supported by both empirical analysis and verified knowledge. It is then
translated into a differentiable constraint on the same variable and operator,
and the update is accepted only when the realized explanation moves toward the
revision without violating predictive guardrails.

Our contributions are threefold. First, we formulate scientific model
refinement as a closed prediction--explanation--knowledge loop rather than a
one-way post-hoc analysis. Second, we introduce typed, minimal explanation
revision, which separates evidence verification from the optimization loss and
prevents an LLM from directly inventing a numerical target. Third, we
instantiate the framework for longitudinal FC-based brain-age prediction and
evaluate prediction, explanation reliability and faithfulness, held-out domain
alignment, targeted revision progress, and off-target explanatory drift.

\begin{figure*}[t]
  \centering
  \includegraphics[width=\textwidth]{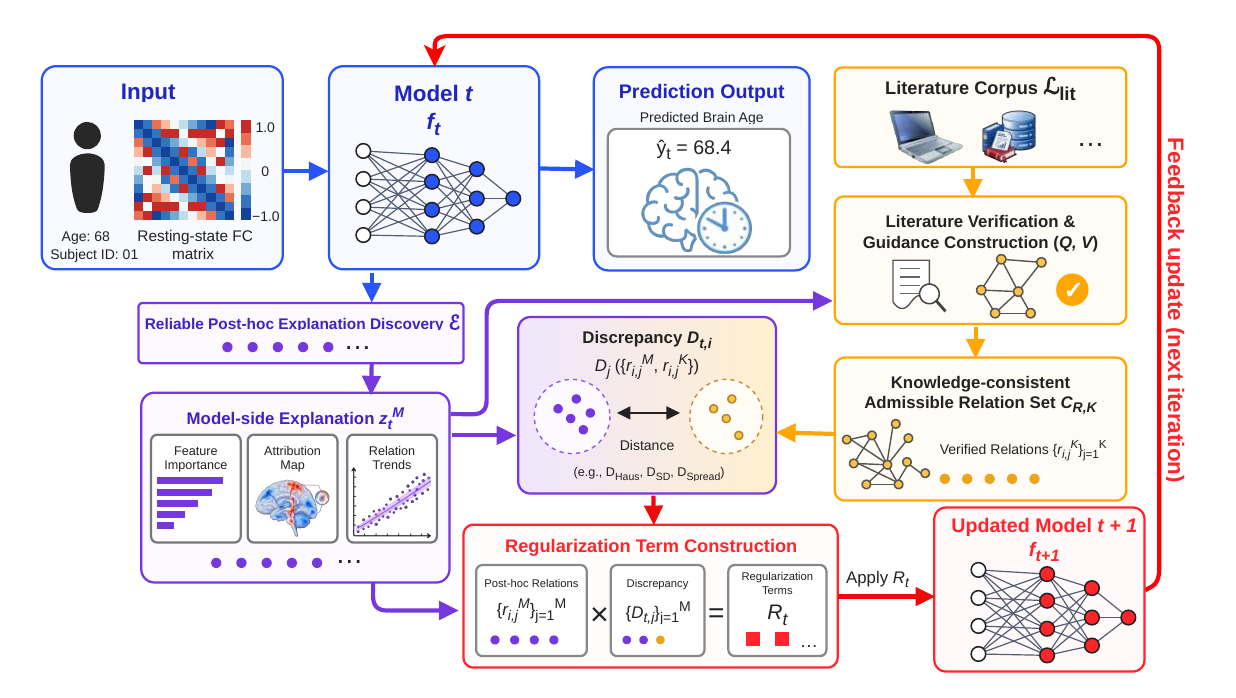
  }
  \caption{End-to-end workflow of \method{}, from reliable post-hoc
  explanation discovery and literature verification to typed minimum-change
  revision, validation-gated model refinement, and iterative re-analysis.}
  \Description{A four-stage loop from prediction to reliable explanation,
  literature verification, minimal explanation revision, and model refinement.}
  \label{fig:framework_overview}
\end{figure*}

\section{Related Work}
\label{sec:related_work}

\subsection{Brain-Age Prediction and Reliable Neuroimaging Explanation}
Functional-connectivity-based age prediction dates to multivariate models of
brain maturity \cite{dosenbach2010prediction}, followed by approaches using
fine-grained FC, interpretable latent networks, multimodal imaging, and graph
architectures \cite{li2018brain,monti2019interpretable,
liem2017predicting,gao2023brain,li2021braingnn,kan2022brain}.
Although these models improve prediction or identify influential regions and
connections, their post-hoc interpretations may be unstable, unfaithful, or
insensitive to model parameters
\cite{adebayo2018sanity,hooker2019benchmark,yeh2019fidelity}.
Unlike prior brain-age studies that mainly report model importance,
\method{} retains reproducible and faithful relations, verifies their
scientific scope, and uses validated revisions to update the predictor.

\subsection{Explanation-, Knowledge-, and Evidence-Guided Learning}
Explanation-guided learning constrains gradients, incorporates interactive
corrections, learns attribution priors, or encodes symbolic and causal
knowledge
\cite{ross2017right,schramowski2020making,rieger2020interpretations,
weinberger2019learning,erion2021improving,xu2018semantic,
kancheti2021matching}. These methods demonstrate that explanations and domain
knowledge can guide training, but generally assume a predefined correction
target. Scientific claim-verification and literature-agent research instead
focuses on retrieving, assessing, and synthesizing scientific evidence
\cite{wadden2020fact,wadden2022multivers,wang2025sciver,
skarlinski2024language,baek2024researchagent}.
In contrast, \method{} discovers candidate relations from the predictor,
constructs scope-aware guidance from verified evidence, and converts the
minimum required explanation revision into a training constraint.

\section{Problem Formulation}
\label{sec:problem_formulation}

We consider a supervised scientific prediction problem over
$\mathcal{D}=\{(\mathbf{x}_i,y_i,m_i)\}_{i=1}^{N}$, where
$\mathbf{x}_i$ is the input, $y_i$ is the prediction target, and $m_i$
contains task-relevant auxiliary information. A predictor $f_{\theta}$
produces $\widehat{y}_i=f_{\theta}(\mathbf{x}_i)$ and is trained by minimizing
\begin{equation}
    \mathcal{L}_{\mathrm{pred}}(\theta)
    =
    \frac{1}{N}
    \sum_{i=1}^{N}
    \ell
    \left(
        f_{\theta}(\mathbf{x}_i),
        y_i
    \right).
    \label{eq:prediction_risk}
\end{equation}
In our brain-age application, $\mathbf{x}_i$ is a functional-connectivity
matrix, $y_i$ is chronological age, and $m_i$ is subject identity.

Beyond accurate prediction, we require the scientific relationships exhibited
by the predictor to be reliably explained and consistent with applicable
verified knowledge. Let
$z_{\theta}=\mathcal{E}(f_{\theta};\mathcal{D})\in\mathcal{Z}$
denote the structured explanation extracted by a validated post-hoc procedure
$\mathcal{E}$, which retains only interpretable patterns that are sufficiently
faithful to the predictor and reproducible across repeated training.

Each retained explanation is treated as a provisional scientific hypothesis.
A query operator $\mathcal{Q}$ translates it into neutral scientific
questions, and a verification operator $\mathcal{V}$ evaluates the questions
against a literature corpus $\mathcal{L}_{\mathrm{lit}}$. The resulting
knowledge record $\kappa_{\theta}$ induces a knowledge-consistent explanation
set $\mathcal{C}_{K}(\kappa_{\theta})\subseteq\mathcal{Z}$.

We formulate the problem as
\begin{equation}
\label{eq:overall_problem}
\begin{aligned}
    \theta^{\star}
    &\in
    \arg\min_{\theta}
    \mathcal{L}_{\mathrm{pred}}(\theta)
    \\
    \mathrm{s.t.}\quad
        z_{\theta^{\star}}
    &=
    \mathcal{E}
    \left(
        f_{\theta^{\star}};
        \mathcal{D}
    \right)
    \in
    \mathcal{C}_{K}
    \left(
        \kappa_{\theta^{\star}}
    \right),
    \\
    \kappa_{\theta^{\star}}
    &=
    \mathcal{V}
    \left(
        \mathcal{Q}
        \left(
            z_{\theta^{\star}}
        \right);
        \mathcal{L}_{\mathrm{lit}}
    \right).
\end{aligned}
\end{equation}

Equation~\eqref{eq:overall_problem} encodes three complementary requirements.
Predictive accuracy is optimized through $\mathcal{L}_{\mathrm{pred}}$;
explainability is operationalized by the validated explanation procedure
$\mathcal{E}$ and the structured space $\mathcal{Z}$; and scientific
reliability is enforced by requiring the resulting explanation to satisfy the
admissible set induced by verified knowledge. The coupled system therefore
seeks an evidence-consistent explanation equilibrium in which the predictor
generates the hypotheses being verified, while the verified knowledge
constrains the explanatory behavior of the predictor.

\section{XAI-Refine}
\label{sec:method}

Directly solving Equation~\eqref{eq:overall_problem} is challenging because
post-hoc analysis, literature verification, knowledge formalization, and model
optimization operate over different representations. XAI-Refine addresses this
challenge through the iterative loop
\begin{equation}
    f_{\theta_t}
    \xrightarrow{\mathcal{E}}
    z_t^{M}
    \xrightarrow{\mathcal{Q},\,\mathcal{V}}
    \kappa_t
    \xrightarrow{\mathcal{P}}
    z_t^{R}
    \xrightarrow{\mathcal{U}}
    f_{\theta_{t+1}}.
    \label{eq:framework_loop}
\end{equation}
The explanation operator $\mathcal{E}$ extracts reliable model-side
explanations (Section~\ref{sec:prediction_to_explanation});
$\mathcal{Q}$ and $\mathcal{V}$ formulate and verify the corresponding
scientific hypotheses (Section~\ref{sec:explanation_to_verification});
$\mathcal{P}$ projects each model-side relation onto its applicable
knowledge-consistent set, and $\mathcal{U}$ updates the predictor to realize
the resulting minimal revisions
(Section~\ref{sec:revision_to_prediction}).

\subsection{Reliable Post-hoc Explanation Discovery}
\label{sec:prediction_to_explanation}

The explanation operator $\mathcal{E}$ characterizes the current predictor
through a collection of post-hoc analyses indexed by $j\in\mathcal{J}$.
Each analysis is specified by a model quantity $v_j$, a measurement operator
$\Phi_j$, and an evaluation scope $\Omega_j$. The operator $\Phi_j$ measures
how the predictor behaves with respect to $v_j$ over $\Omega_j$, producing an
explanatory relation in $\mathcal{Z}_j$. Depending on the analysis, this
relation may describe prediction behavior, feature response, longitudinal
consistency, attribution structure, or representation geometry.

Because an explanation obtained from a single training run may reflect a
particular initialization, we train the predictor under $S$ random seeds and
consolidate the resulting measurements:
\begin{equation}
    r_{t,j}^{M}
    =
    \operatorname{Agg}_{j}
    \left(
        \left\{
            \Phi_j
            \left(
                f_{\theta_t}^{(s)};
                v_j,\Omega_j
            \right)
        \right\}_{s=1}^{S}
    \right).
    \label{eq:model_explanation_aggregation}
\end{equation}
Here, $r_{t,j}^{M}\in\mathcal{Z}_j$ denotes the explanatory relation exhibited
by the current predictor. The aggregation is source-preserving in the sense that all seed-wise
measurements and their provenance are retained. It may summarize compatible
measurements into a common relation, but it cannot change the analyzed model
quantity, measurement operator, or evaluation scope, nor can it introduce a
direction or scientific interpretation unsupported by the source
measurements. We evaluate the reproducibility of $r_{t,j}^{M}$ using support, coverage, and
cross-seed stability:
\begin{equation}
    \rho_{t,j}
    =
    \min
    \left\{
        \rho_{t,j}^{\mathrm{sup}},
        \rho_{t,j}^{\mathrm{cov}},
        \rho_{t,j}^{\mathrm{stab}}
    \right\}.
    \label{eq:explanation_reliability}
\end{equation}
Support measures how consistently the relation is reproduced among evaluable
runs, coverage measures how broadly it can be assessed, and stability measures
numerical or structural agreement across seeds. 

Cross-seed reproducibility alone does not guarantee that a post-hoc result
accurately reflects the predictor. We therefore additionally compute a
faithfulness score $\operatorname{Fid}_{t,j}$ using a validation procedure
appropriate to analysis $j$, such as perturbation response, model
randomization, input randomization, or direct agreement with observed model
behavior.
The structured model-side explanation produced by $\mathcal{E}$ is
\begin{equation}
\label{eq:model_explanation_state}
\begin{aligned}
    z_t^{M}
    &=
    \mathcal{E}
    \left(
        f_{\theta_t};
        \mathcal{D}_{\mathrm{tr}}
    \right)
    =
    \left\{
        e_{t,j}^{M}
        \,\middle|\,
        j\in\mathcal{J},
        \;
        \rho_{t,j}\geq\tau_{\rho},
        \;
        \operatorname{Fid}_{t,j}\geq\tau_F
    \right\},
    \\
    e_{t,j}^{M}
    &=
    \left(
        v_j,
        \Phi_j,
        \Omega_j,
        r_{t,j}^{M},
        \rho_{t,j},
        \operatorname{Fid}_{t,j}
    \right).
\end{aligned}
\end{equation}
Here, $z_t^{M}$ denotes the collection of model-valid explanation records,
whereas $r_{t,j}^{M}\in\mathcal{Z}_j$ denotes the relation extracted by
analysis $j$. Each retained explanation preserves the model quantity $v_j$,
the measurement operator $\Phi_j$, the evaluation scope $\Omega_j$, the
measured relation $r_{t,j}^{M}$, its cross-seed reproducibility
$\rho_{t,j}$, and its faithfulness score $\operatorname{Fid}_{t,j}$.
Only explanations that satisfy both validation criteria are passed to
literature verification. Explanations that fail either criterion remain in the
audit trail but do not generate scientific queries or model-refinement
constraints. Retaining $v_j$, $\Phi_j$, and $\Omega_j$ ensures that any
subsequent revision is defined and evaluated on the same explanatory quantity
used to diagnose the original model behavior.
Detailed analysis and validation protocols are
provided in Appendix~\ref{app:module1_details}.

\subsection{Literature-Grounded Guidance Construction}
\label{sec:explanation_to_verification}

A retained explanation characterizes the behavior of the current predictor,
but does not by itself constitute verified scientific knowledge. For each
$e_{t,j}^{M}\in z_t^{M}$, the query operator constructs a neutral scientific
question, which is then assessed against the literature:
\begin{equation}
    q_{t,j}
    =
    \mathcal{Q}
    \left(
        e_{t,j}^{M}
    \right),
    \qquad
    \kappa_{t,j}
    =
    \mathcal{V}
    \left(
        q_{t,j};
        \mathcal{L}_{\mathrm{lit}}
    \right).
    \label{eq:literature_verification}
\end{equation}
The query preserves the scientific variable, relation type, population,
measurement setting, and scope represented by
$(v_j,\Phi_j,\Omega_j)$, while withholding the model-observed direction,
magnitude, and reliability whenever revealing them could bias retrieval toward
confirmation. For example, suppose that $v_j$ denotes the age-related response
of a hippocampal connectivity pattern, $\Phi_j$ measures its standardized slope
with respect to chronological age, and the current model reports
$r_{t,j}^{M}=+0.12$ among participants aged $60$--$80$. The query asks how the
same connectivity pattern varies with age in a comparable population without
revealing the positive model-observed slope.

To construct the knowledge-side record, the verification operator
$\mathcal{V}$ performs literature retrieval, bibliographic validation,
evidence extraction, and scope assessment. For each query $q_{t,j}$, it
returns a structured, provenance-bearing knowledge record:
\begin{equation}
\label{eq:verified_knowledge_state}
\begin{aligned}
    \kappa_{t,j}
    &=
    \left(
        v_j,
        \Omega_{t,j}^{K},
        r_{t,j}^{K},
        c_{t,j},
        s_{t,j},
        \mathcal{B}_{t,j}
    \right),
    \\
    \kappa_t
    &=
    \left\{
        \kappa_{t,j}
        \,\middle|\,
        e_{t,j}^{M}\in z_t^{M}
    \right\}.
\end{aligned}
\end{equation}
Here, $\Omega_{t,j}^{K}$ specifies the scope supported by the retrieved
evidence, $r_{t,j}^{K}$ records the corresponding knowledge-side relation,
$c_{t,j}$ measures evidence confidence, and $s_{t,j}$ records the evidence
status. The set $\mathcal{B}_{t,j}$ retains the verified publications,
evidence excerpts, and bibliographic identifiers supporting the record.
Continuing the example, the literature may support a negative age-related
relation, $r_{t,j}^{K}<0$, with confidence $c_{t,j}=0.80$ among participants
aged $55$--$85$.

The verified knowledge record is next aligned with its originating model
explanation. Let
\begin{equation}
    \Omega_{t,j}^{\cap}
    =
    \operatorname{Align}
    \left(
        \Omega_j,
        \Omega_{t,j}^{K}
    \right)
\end{equation}
denote the population, anatomical, measurement, and experimental scope over
which the model-side and knowledge-side relations are comparable.

A deterministic compiler then maps the verified relation into the same typed
explanation space as the originating model relation:
\begin{equation}
\label{eq:knowledge_set_compilation}
    \mathcal{C}_{K,t,j}
    =
    \operatorname{Compile}
    \left(
        \kappa_{t,j};
        \mathcal{Z}_j,
        \Omega_{t,j}^{\cap}
    \right)
    \subseteq
    \mathcal{Z}_j.
\end{equation}

The compiled set describes which values of the explanatory relation are
consistent with the applicable verified evidence. For example, verified
non-positive direction may induce
$\mathcal{C}_{K,t,j}=(-\infty,0]$, whereas evidence supporting a negative
margin may induce
$\mathcal{C}_{K,t,j}=(-\infty,-m_{t,j}]$ for $m_{t,j}>0$.
Interval, ordering, monotonicity, distributional, and geometric evidence
produce corresponding admissible sets in their associated explanation spaces.

The compiler accounts for evidence uncertainty and specificity. Strong and
directly applicable evidence may produce a narrow admissible set, while broad
or mixed evidence produces a less restrictive set. When no applicable evidence
is available, the compiler returns
$\mathcal{C}_{K,t,j}=\mathcal{Z}_j$, which produces no literature-driven
revision. If the aligned scope is empty or the evidence cannot be represented
as a nonempty closed admissible set, the record is deferred rather than
translated into a training constraint.

The resulting guidance record is
\begin{equation}
\label{eq:knowledge_guidance_record}
    \gamma_{t,j}
    =
    \left(
        v_j,
        \Phi_j,
        \Omega_{t,j}^{\cap},
        \mathcal{C}_{K,t,j},
        w_{t,j}
    \right),
    \qquad
    \gamma_t
    =
    \left\{
        \gamma_{t,j}
        \,\middle|\,
        e_{t,j}^{M}\in z_t^{M}
    \right\}.
\end{equation}

Here, $w_{t,j}\in[0,1]$ summarizes the reliability of the complete guidance
record based on the reproducibility and faithfulness of the model explanation,
the confidence and applicability of the verified evidence, and the
compatibility of their scopes. It determines whether a revision is eligible
for execution, but it is not itself a scientific target.

Importantly, the literature-verification agent does not directly output an
instruction such as ``strengthen'' or ``reverse.'' It produces a structured
knowledge record, and the compiler converts that record into an admissible set.
The actual direction and magnitude of revision are determined subsequently by
projecting the current model relation onto this set.
Further details of scope alignment, relation comparison, and
guidance construction are provided in
Appendix~\ref{app:module2_details}.

\subsection{Explanation-Guided Model Refinement}
\label{sec:revision_to_prediction}

The verified knowledge state specifies which explanatory relations are
admissible, but it does not provide a complete replacement for the current
model explanation. We therefore revise each retained relation through a
minimum-change projection in its own typed explanation space.

For explanation analysis $j$, the revised relation is
\begin{equation}
\label{eq:componentwise_revision}
    r_{t,j}^{R}
    =
    \operatorname{Proj}_{\mathcal{C}_{K,t,j}}
    \left(
        r_{t,j}^{M}
    \right)
    \in
    \arg\min_{
        r\in\mathcal{C}_{K,t,j}
    }
    D_j
    \left(
        r,
        r_{t,j}^{M}
    \right),
\end{equation}
where $D_j$ is defined according to the representation of the explanatory
relation in $\mathcal{Z}_j$. Scalar, vector, ranking, distributional, and
geometric relations may therefore use different but type-compatible
discrepancy measures.

For executable revisions, the compiler returns a nonempty closed admissible
set, ensuring that the projection is well defined. When convexity applies, the
set and discrepancy may additionally be chosen so that the projection is
unique. Records that cannot be represented by such an executable set are
deferred.

The complete revised explanation state is
\begin{equation}
\label{eq:revised_explanation_state}
    z_t^{R}
    =
    \left\{
        e_{t,j}^{R}
        \,\middle|\,
        e_{t,j}^{M}\in z_t^{M}
    \right\},
    \qquad
    e_{t,j}^{R}
    =
    \left(
        v_j,
        \Phi_j,
        \Omega_{t,j}^{\cap},
        r_{t,j}^{R},
        w_{t,j}
    \right).
\end{equation}

The projection changes only the part of the explanation required for knowledge
consistency. If
$r_{t,j}^{M}\in\mathcal{C}_{K,t,j}$, then
$r_{t,j}^{R}=r_{t,j}^{M}$. Otherwise, the projection returns the nearest
admissible relation rather than copying a numerical effect magnitude from the
literature.

For example, suppose that the model exhibits a positive scalar relation
$r_{t,j}^{M}>0$, whereas the applicable evidence supports a non-positive
relation. The compiler defines
\begin{equation}
    \mathcal{C}_{K,t,j}
    =
    (-\infty,0],
\end{equation}
and squared-distance projection gives
\begin{equation}
    r_{t,j}^{R}
    =
    \min
    \left(
        r_{t,j}^{M},
        0
    \right).
\end{equation}
If the evidence supports a negative margin $m_{t,j}>0$, the corresponding set
is $(-\infty,-m_{t,j}]$. Thus, verified knowledge determines the admissible
region, while projection determines the smallest model-specific revision.

Not every projected difference warrants model retraining. We define the
actionable revision set as
\begin{equation}
\label{eq:actionable_revisions}
    \mathcal{A}_t
    =
    \left\{
        j
        \,\middle|\,
        D_j
        \left(
            r_{t,j}^{M},
            r_{t,j}^{R}
        \right)
        >
        \varepsilon_{\mathrm{eq}},
        \;
        w_{t,j}
        \geq
        \tau_w,
        \;
        \Omega_{t,j}^{\cap}
        \neq
        \varnothing
    \right\}.
\end{equation}
Hence, a revision is executed only when it is non-negligible, sufficiently
well supported, and applicable to a nonempty aligned scope.

For each $j\in\mathcal{A}_t$, the candidate predictor is evaluated using the
same explanatory variable and measurement operator that produced the original
model-side explanation:
\begin{equation}
\label{eq:realization_discrepancy}
    \delta_{t,j}
    \left(
        \theta
    \right)
    =
    D_j
    \left(
        \Phi_j
        \left(
            f_{\theta};
            v_j,
            \Omega_{t,j}^{\cap}
        \right),
        r_{t,j}^{R}
    \right).
\end{equation}

Conceptually, the model update seeks the most accurate predictor that realizes
the actionable revised explanations:
\begin{equation}
\label{eq:constrained_model_update}
\begin{aligned}
    \widetilde{\theta}_{t+1}
    \in
    \arg\min_{\theta}\quad
    &
    \mathcal{L}_{\mathrm{pred}}
    \left(
        \theta;
        \mathcal{D}_{\mathrm{tr}}
    \right)
    \\
    \mathrm{s.t.}\quad
    &
    \delta_{t,j}
    \left(
        \theta
    \right)
    \leq
    \varepsilon_{R,j},
    \qquad
    j\in\mathcal{A}_t.
\end{aligned}
\end{equation}

For differentiable implementation, we optimize its violation-penalty form
\begin{equation}
\label{eq:constraint_violation_objective}
    \widetilde{\theta}_{t+1}
    \in
    \arg\min_{\theta}
    \left[
        \mathcal{L}_{\mathrm{pred}}
        \left(
            \theta;
            \mathcal{D}_{\mathrm{tr}}
        \right)
        +
        \lambda
        \sum_{j\in\mathcal{A}_t}
        \alpha
        \left(
            w_{t,j}
        \right)
        \left[
            \delta_{t,j}
            \left(
                \theta
            \right)
            -
            \varepsilon_{R,j}
        \right]_{+}
    \right],
\end{equation}
where $[a]_{+}=\max(0,a)$ and
$\alpha(w_{t,j})$ maps scientific guidance confidence to a calibrated
optimization budget. When $\Phi_j$ is not differentiable, training uses a
compatible differentiable surrogate, while the original operator $\Phi_j$ is
reapplied after optimization to evaluate whether the intended relation was
actually realized.

The candidate model is evaluated on $\mathcal{D}_{\mathrm{val}}$ and promoted
only when three conditions are satisfied: predictive degradation remains
within a predefined guardrail, the targeted explanation discrepancies decrease
relative to the parent model, and non-target explanatory behavior remains
within an allowable drift threshold. Formally,
\begin{equation}
\label{eq:model_promotion}
    f_{\theta_{t+1}}
    =
    \begin{cases}
        f_{\widetilde{\theta}_{t+1}},
        &
        \text{if the validation promotion criteria are satisfied},
        \\
        f_{\theta_t},
        &
        \text{otherwise}.
    \end{cases}
\end{equation}

After promotion, the updated predictor is reanalyzed using the same explanation
and verification procedures, closing the loop.

For the set of eligible and applicable explanations
\begin{equation}
    \mathcal{J}_t^{\mathrm{elig}}
    =
    \left\{
        j
        \,\middle|\,
        w_{t,j}\geq\tau_w,
        \;
        \Omega_{t,j}^{\cap}\neq\varnothing
    \right\},
\end{equation}
an evidence-consistent explanation equilibrium is reached when
\begin{equation}
\label{eq:equilibrium_condition}
    r_{t,j}^{R}
    =
    r_{t,j}^{M},
    \qquad
    \forall
    j\in
    \mathcal{J}_t^{\mathrm{elig}}.
\end{equation}
At this point, every eligible model relation already belongs to its applicable
knowledge-consistent set and no knowledge-driven revision is required.

The absence of an actionable revision does not always establish equilibrium.
It may instead reflect insufficient evidence, low confidence, an empty aligned
scope, or the absence of an executable constraint. The loop therefore
distinguishes evidence-consistent equilibrium from the weaker condition that
no actionable update exists under the current evidence and operation library.
It terminates when equilibrium is reached, when no feasible promoted update is
found for a predefined number of consecutive iterations, or when the maximum
number of iterations is reached.

\section{Experiments}
\label{sec:experiments}

The experiments are designed around six questions. \textbf{RQ1} asks whether
\method{} improves the overall scientific quality of the final predictor when
prediction, explanation, and domain alignment are evaluated jointly.
\textbf{RQ2} tests whether explanations entering the loop are reproducible and
faithful. \textbf{RQ3} evaluates literature verification and the construction
of minimal revised explanations. \textbf{RQ4} tests whether revision-guided
training causes target-specific explanatory movement rather than generic
retraining drift. \textbf{RQ5} examines iterative progress and generalization
to unseen subjects, model runs, and literature. \textbf{RQ6} studies ablations,
response-probing efficiency, and sensitivity.

\subsection{Experimental Setup}
\label{sec:experimental_setup}

\paragraph{Data and predictor.}
We use resting-state fMRI scans from 613 participants in the controlled-access longitudinal cohort of the Emory Healthy Brain Study (EHBS). Each scan is represented by an
$81\times81$  functional-connectivity matrix constructed
over 81 brain regions and is paired with chronological age as the prediction
target. All partitions are subject-disjoint, ensuring that scans from the same
participant never appear in different splits. Subject identity is used only to
construct longitudinal pairs and same-subject analyses, and is not provided to
the age predictor. Table~\ref{tab:data_split} summarizes the split statistics.

\begin{table}[t]
    \centering
    \caption{Dataset composition and longitudinal coverage for the
subject-disjoint training, validation, and test splits.}

    \label{tab:data_split}
    \scriptsize
    \setlength{\tabcolsep}{3.2pt}
    \renewcommand{\arraystretch}{1.08}
    \resizebox{\columnwidth}{!}{
    \begin{tabular}{lcccc}
        \toprule
        & Train & Validation & Test & Total \\
        \midrule
        Subjects (scans)
        & 429 (906)
        & 92 (198)
        & 92 (193)
        & 613 (1,297) \\
        Age, mean$\pm$SD
        & 66.11$\pm$6.73
        & 66.39$\pm$6.52
        & 66.55$\pm$6.34
        & 66.22$\pm$6.64 \\
        Age range
        & 50.27--82.10
        & 51.50--81.80
        & 49.80--83.90
        & 49.80--83.90 \\
        Longitudinal subjects
        & 415 (96.7\%)
        & 89 (96.7\%)
        & 90 (97.8\%)
        & 594 (96.9\%) \\
        \bottomrule
    \end{tabular}
    }
\end{table}

The primary predictor contains two TransformerConv layers followed by
graph-level pooling and a scalar regression head. Discovery models are trained
under five random seeds, while additional seeds are reserved for explanation
confirmation. The training set is used for model fitting and explanation
discovery, the validation set for response probing and model promotion, and the
test set only after the complete refinement trajectory has been fixed.

\paragraph{Compared methods.}
We compare \method{} with representative predictive baselines from three
categories. Classical vector-based methods include Ridge
regression~\cite{hoerl2000ridge}, RBF-SVR~\cite{lee2009advances}, Random
Forest~\cite{breiman2001random}, Elastic Net~\cite{zou2005regularization}, and
XGBoost~\cite{chen2016xgboost}. General neural and graph baselines include an
MLP~\cite{rumelhart1986learning}, GCN~\cite{kipf2016semi},
GAT~\cite{Velickovic2017GraphAN}, and the prediction-only TransformerConv
backbone $f_0$~\cite{Shi2020MaskedLP}. Brain-network-specific baselines include
BrainNetCNN~\cite{Kawahara2017BrainNetCNNCN},
BrainGNN~\cite{li2021braingnn}, Brain Network
Transformer~\cite{kan2022brain}, the rs-fMRI brain-age
GNN~\cite{gao2023brain}, the FC foundation
model~\cite{Wang2025AGT}, and
BioBGT~\cite{peng2025biologicallyplausiblebraingraph}.

We further compare \method{} with controlled variants built on the same
prediction-only backbone $f_0$: \emph{prediction only}, with no
explanation-guided refinement; \emph{random-target constraints}, which apply
the proposed constraints to randomly selected targets;
\emph{opposite-direction constraints}, which reverse the prescribed response
directions; \emph{direct evidence-to-loss}, which converts verified evidence
directly into training objectives without typed revision; \emph{one-shot typed
revision}, which derives guidance from $f_0$ once without subsequent
re-analysis; and \emph{all eligible revisions}, which applies all actionable
constraints without response probing or filtering. The complete iterative
\method{} loop additionally re-analyzes each updated predictor and admits
revisions through response probing and validation. All variants use the same
subject-disjoint splits, matched seeds, optimization settings, and training
budgets where applicable. Further details are provided in
Appendix~\ref{app:compared_methods}.

\subsection{Evaluation Metrics}
\label{sec:evaluation_metrics}

We evaluate four dimensions using eight metrics. MAE and PCC measure
prediction accuracy and age-order consistency, respectively. Confirmation
measures whether discovered explanation claims reappear on held-out data,
whereas Faithfulness measures whether perturbing a claimed feature changes the
prediction in the claimed direction. CSR measures how many stable explanation
claims receive supporting literature evidence, while AEA measures whether
better-supported features receive larger and higher-ranked attributions.
Progress measures how many intended revision targets improve after model
refinement, whereas Drift measures unintended changes in non-target model
behavior. Detailed definitions are provided in
Appendix~\ref{app:evaluation_metrics}.

\subsection{RQ1: Overall Scientific-Model Comparison}
\label{sec:overall_results}

Table~\ref{tab:overall_results} compares predictive baselines and controlled
refinement strategies across prediction, explanation quality, domain
alignment, and explanatory revision. Within each panel, bold and underlined
values denote the best and second-best results, respectively.

\begin{table*}[!t]
\centering
\scriptsize
\caption{Overall comparison of predictive baselines and matched refinement
strategies across prediction, explanation reliability, domain alignment, and
controlled revision. Panel~A reports independently trained predictive models,
whereas Panel~B reports refinement strategies initialized from the same parent
predictor. Dashes indicate metrics that are undefined without a matched
parent--child refinement transition.}
\label{tab:overall_results}
\setlength{\tabcolsep}{3.2pt}
\renewcommand{\arraystretch}{1.08}
\begin{tabular}{ll|cc|cc|cc|cc}

\toprule
&
&
\multicolumn{2}{c|}{Prediction}
&
\multicolumn{2}{c|}{Explanation}
&
\multicolumn{2}{c|}{Domain Alignment}
&
\multicolumn{2}{c}{Controlled Revision}
\\
Category
&
Method
&
MAE $\downarrow$
&
PCC $\uparrow$
&
Confirm. $\uparrow$
&
Faith. $\uparrow$
&
CSR $\uparrow$
&
AEA $\uparrow$
&
Progress $\uparrow$
&
Drift $\downarrow$
\\
\midrule

\multicolumn{10}{l}{\textbf{Panel A: Predictive baselines}}
\\
\midrule

Classical
&
Ridge
&
5.8478 $\pm$ 0.0000
&
0.4420 $\pm$ 0.0000
&
\textbf{1.0000 $\pm$ 0.0000}
&
0.9432 $\pm$ 0.0000
&
0.2418
&
0.3819 $\pm$ 0.0000
&
--
&
--
\\

Classical
&
SVR-RBF
&
4.6935 $\pm$ 0.0000
&
\underline{0.5075 $\pm$ 0.0000}
&
\textbf{1.0000 $\pm$ 0.0000}
&
0.9199 $\pm$ 0.0000
&
0.2665
&
\underline{0.4026 $\pm$ 0.0000}
&
--
&
--
\\

Classical
&
Random Forest
&
4.7320 $\pm$ 0.0371
&
0.4695 $\pm$ 0.0136
&
\underline{0.9939 $\pm$ 0.0136}
&
0.3412 $\pm$ 0.0134
&
0.2737
&
0.3313 $\pm$ 0.0054
&
--
&
--
\\

Classical
&
Elastic Net
&
5.9714 $\pm$ 0.0000
&
0.4309 $\pm$ 0.0000
&
\textbf{1.0000 $\pm$ 0.0000}
&
\underline{0.9889 $\pm$ 0.0000}
&
0.2418
&
0.2908 $\pm$ 0.0000
&
--
&
--
\\

Classical
&
XGBoost
&
\underline{4.5104 $\pm$ 0.0395}
&
0.4881 $\pm$ 0.0079
&
\textbf{1.0000 $\pm$ 0.0000}
&
0.1260 $\pm$ 0.0094
&
0.2405
&
0.2695 $\pm$ 0.0063
&
--
&
--
\\

General neural/graph
&
MLP
&
4.5213 $\pm$ 0.2140
&
\textbf{0.5116 $\pm$ 0.0301}
&
0.7394 $\pm$ 0.2274
&
0.8216 $\pm$ 0.0606
&
0.2717
&
0.3383 $\pm$ 0.0100
&
--
&
--
\\

General neural/graph
&
GCN
&
4.6721 $\pm$ 0.1870
&
0.4373 $\pm$ 0.0375
&
0.6424 $\pm$ 0.1412
&
0.9019 $\pm$ 0.0124
&
0.2985
&
0.3834 $\pm$ 0.0179
&
--
&
--
\\

General neural/graph
&
GAT
&
4.6631 $\pm$ 0.2986
&
0.4824 $\pm$ 0.0636
&
0.6065 $\pm$ 0.1954
&
0.8266 $\pm$ 0.0273
&
0.3109
&
0.3600 $\pm$ 0.0408
&
--
&
--
\\

General neural/graph
&
TransformerConv baseline
&
\textbf{4.5023 $\pm$ 0.0493}
&
0.4867 $\pm$ 0.0188
&
0.4062 $\pm$ 0.1169
&
0.8351 $\pm$ 0.0229
&
\textbf{0.3831}
&
0.3958 $\pm$ 0.0394
&
--
&
--
\\

Brain-specific
&
BrainNetCNN
&
4.6257 $\pm$ 0.0617
&
0.4717 $\pm$ 0.0276
&
0.6848 $\pm$ 0.0899
&
0.8193 $\pm$ 0.0323
&
0.1967
&
0.3067 $\pm$ 0.0072
&
--
&
--
\\

Brain-specific
&
BrainGNN
&
4.9356 $\pm$ 0.1412
&
0.3487 $\pm$ 0.0511
&
0.3630 $\pm$ 0.1896
&
0.6598 $\pm$ 0.0645
&
0.3230
&
0.3336 $\pm$ 0.0713
&
--
&
--
\\

Brain-specific
&
Brain Network Transformer
&
4.6434 $\pm$ 0.1129
&
0.4751 $\pm$ 0.0197
&
0.8485 $\pm$ 0.0606
&
\textbf{0.9927 $\pm$ 0.0041}
&
0.2711
&
0.3651 $\pm$ 0.0094
&
--
&
--
\\

Brain-specific
&
rs-fMRI Brain-Age GNN
&
4.9827 $\pm$ 0.1739
&
0.3789 $\pm$ 0.0421
&
0.4606 $\pm$ 0.0919
&
0.8251 $\pm$ 0.0515
&
0.3172
&
0.3922 $\pm$ 0.0271
&
--
&
--
\\

Brain-specific
&
FC foundation model
&
5.1111 $\pm$ 0.2873
&
0.3211 $\pm$ 0.0206
&
0.0000 $\pm$ 0.0000
&
0.0000 $\pm$ 0.0000
&
0.3128
&
0.2826 $\pm$ 0.0492
&
--
&
--
\\

Brain-specific
&
BioBGT
&
4.5797 $\pm$ 0.1065
&
0.4871 $\pm$ 0.0204
&
0.6182 $\pm$ 0.1569
&
0.9585 $\pm$ 0.0026
&
\underline{0.3239}
&
\textbf{0.4186 $\pm$ 0.0604}
&
--
&
--
\\

\midrule
\multicolumn{10}{l}{\textbf{Panel B: Controlled refinement}}
\\
\midrule

Parent
&
Prediction only $f_0$
&
4.5360 $\pm$ 0.1127
&
0.4788 $\pm$ 0.0174
&
0.4000 $\pm$ 0.1412
&
0.7967 $\pm$ 0.0486
&
0.3166
&
0.3646 $\pm$ 0.0484
&
--
&
--
\\

Refinement
&
Random-target constraints
&
\underline{4.4600 $\pm$ 0.1191}
&
\underline{0.5017 $\pm$ 0.0292}
&
0.5437 $\pm$ 0.1373
&
0.8081 $\pm$ 0.0133
&
0.3354
&
0.3607 $\pm$ 0.0495
&
0.5333 $\pm$ 0.2767
&
\textbf{0.0000 $\pm$ 0.0000}
\\

Refinement
&
Opposite-direction constraints
&
4.6206 $\pm$ 0.0931
&
0.4645 $\pm$ 0.0231
&
0.4714 $\pm$ 0.1273
&
0.7328 $\pm$ 0.0321
&
0.3481
&
0.3754 $\pm$ 0.0553
&
0.5111 $\pm$ 0.1685
&
0.0571 $\pm$ 0.0782
\\

Refinement
&
Direct evidence-to-loss
&
4.5956 $\pm$ 0.1787
&
0.4798 $\pm$ 0.0202
&
\textbf{0.6875 $\pm$ 0.1362}
&
\underline{0.8153 $\pm$ 0.0358}
&
0.3364
&
0.3787 $\pm$ 0.0365
&
0.5778 $\pm$ 0.1648
&
0.0571 $\pm$ 0.1278
\\

Refinement
&
One-shot typed revision
&
4.4943 $\pm$ 0.1901
&
0.4910 $\pm$ 0.0381
&
0.4533 $\pm$ 0.0767
&
0.7623 $\pm$ 0.0431
&
\textbf{0.4933}
&
\underline{0.4036 $\pm$ 0.0545}
&
0.6000 $\pm$ 0.3566
&
\underline{0.0286 $\pm$ 0.0639}
\\

Refinement
&
All eligible revisions without probing
&
4.4943 $\pm$ 0.1059
&
0.4950 $\pm$ 0.0289
&
0.5875 $\pm$ 0.1046
&
0.7671 $\pm$ 0.0351
&
0.2829
&
0.3800 $\pm$ 0.0379
&
\underline{0.6667 $\pm$ 0.1361}
&
\underline{0.0286 $\pm$ 0.0639}
\\

Proposed
&
\method{}
&
\textbf{4.3813 $\pm$ 0.0258}
&
\textbf{0.5274 $\pm$ 0.0131}
&
\underline{0.6500 $\pm$ 0.0601}
&
\textbf{0.8455 $\pm$ 0.0084}
&
\underline{0.4794}
&
\textbf{0.5722 $\pm$ 0.0112}
&
\textbf{0.8444 $\pm$ 0.0609}
&
\underline{0.0286 $\pm$ 0.0639}
\\

\bottomrule
\end{tabular}
\end{table*}

Panel A shows that no independently trained predictive architecture dominates
all evaluation dimensions. The TransformerConv baseline achieves the lowest
MAE and the highest CSR among predictive baselines, while the MLP obtains the
highest PCC. Brain Network Transformer achieves the highest Faithfulness, and
BioBGT obtains the strongest AEA. These differences illustrate that predictive
accuracy, explanation reproducibility, faithfulness, and literature alignment
capture distinct properties of a scientific predictor.

In Panel B, \method{} achieves the best MAE and PCC and improves Faithfulness, AEA, and target Progress relative to the prediction-only parent. Its Confirmation and CSR are second only to direct
evidence-to-loss and one-shot typed revision, respectively. Direct
evidence-to-loss produces the highest Confirmation but weaker prediction,
domain alignment, and target progress, while one-shot revision achieves the
highest CSR but does not match the iterative loop in AEA or Progress. Applying
all eligible revisions without probing improves Progress but remains below the
complete loop, supporting the role of response-based selection and iterative
re-analysis.

Relative to $f_0$, the final \method{} model reduces MAE from
$4.5360$ to $4.3813$, increases PCC from $0.4788$ to $0.5274$, raises
Confirmation from $0.4000$ to $0.6500$, and improves AEA from $0.3646$ to
$0.5722$. It realizes $84.44\%$ of the prespecified target directions while
producing almost no detected off-target change beyond prediction-only retraining
variation.

\subsection{RQ2--RQ3: Explanation, Verification, and Revision Validity}
\label{sec:component_validity}

We evaluate explanation discovery and literature-grounded revision separately.
RQ2 compares single-run discovery, multi-run aggregation, and the full reliable
discovery protocol. All explanation claims are frozen before evaluation on
held-out subjects and independently trained models. RQ3 conducts a sequential
single-reviewer audit of 37 literature-grounding records, where the evidence is
assessed before the model claim and proposed revision are revealed.

\begin{table}[t]
\centering
\caption{Comparison of explanation-discovery protocols under single-run
analysis, multi-run aggregation, and full reproducibility--faithfulness
filtering.}
\label{tab:explanation_ablation}

\scriptsize
\renewcommand{\arraystretch}{1.08}

\begin{tabular*}{\columnwidth}{
    @{\extracolsep{\fill}}lccc@{}
}
\toprule
Protocol
& Claims
& Confirm. $\uparrow$
& Faith. $\uparrow$
\\
\midrule

Single run
& 33
& 0.1091 $\pm$ 0.0664
& 0.6091 $\pm$ 0.0105
\\

Multi-run aggregation
& 32
& \underline{0.4500 $\pm$ 0.1351}
& \underline{0.7726 $\pm$ 0.0518}
\\

Full reliable discovery
& 29
& \textbf{0.6500 $\pm$ 0.0601}
& \textbf{0.8455 $\pm$ 0.0084}
\\

\bottomrule
\end{tabular*}
\end{table}
\begin{table}[t]
\centering
\caption{Single-reviewer audit of evidence grounding, precise abstention,
and typed revision validity.}
\label{tab:verification_audit}

\scriptsize
\renewcommand{\arraystretch}{1.10}

\begin{tabular*}{\columnwidth}{
    @{\extracolsep{\fill}}lcc@{}
}
\toprule
Audit metric
& Agreement
& Rate
\\
\midrule

Evidence-status agreement
& 31/37
& 83.8\%
\\

Strict scope agreement
& 23/33
& 69.7\%
\\

Precise-abstention agreement
& 33/37
& 89.2\%
\\

Admissible revisions
& 32/37
& 86.5\%
\\

Variable preservation
& 37/37
& 100.0\%
\\

Operator preservation
& 36/37
& 97.3\%
\\

Revision-scope preservation
& 35/37
& 94.6\%
\\

Minimum-revision correctness
& 32/37
& 86.5\%
\\

\bottomrule
\end{tabular*}
\end{table}
As shown in Table~\ref{tab:explanation_ablation}, multi-run aggregation
substantially improves both held-out confirmation and perturbation
faithfulness over single-run discovery. The full reliability protocol further
filters three claims while increasing Confirmation to $0.6500$ and
Faithfulness to $0.8455$. This indicates that cross-run aggregation and
faithfulness filtering jointly improve both the reproducibility and behavioral
validity of the retained explanations.

Table~\ref{tab:verification_audit} shows strong reviewer agreement for evidence
status, precise abstention, and revision validity. Variable, operator, and
revision-scope preservation all exceed $94\%$, indicating that the generated
revisions generally modify the intended explanatory quantity. Strict scope
agreement is lower at $69.7\%$, with the main errors arising when broad or
connection-family evidence is compiled into an overly specific strengthening
constraint. Detailed protocols and error analyses are provided in
Appendix~\ref{app:component_validity}.

\subsection{RQ4: Targeted Effect and Update Specificity}
\label{sec:refinement_results}

We evaluate whether repeated knowledge-guided updates move the intended model
behaviors toward their targets while preserving non-target behavior and
predictive performance. All refinement strategies follow the same sequence of
six scientific revisions. Random-target and opposite-direction controls test
whether improvement arises from the scientific content of the constraints
rather than from adding regularization alone.

\begin{table*}[!t]
\centering
\scriptsize
\caption{Targeted refinement performance across six matched rounds.
NTGC and TIP quantify movement toward the intended targets; NOTC and MOCR
quantify the magnitude and prevalence of off-target change; TRS measures
update specificity; PPR measures predictive preservation; and DPR measures
validation-to-test persistence. }
\label{tab:target_results}
\setlength{\tabcolsep}{3.3pt}
\renewcommand{\arraystretch}{1.08}
\resizebox{\textwidth}{!}{
\begin{tabular}{lccccccc}
\toprule
Strategy
& NTGC $\uparrow$
& TIP $\uparrow$
& NOTC $\downarrow$
& MOCR $\downarrow$
& TRS $\uparrow$
& PPR $\uparrow$
& DPR $\uparrow$
\\
\midrule

backbone $f_0$
& 0.0000 $\pm$ 0.0000
& 0.0000 $\pm$ 0.0000
& 0.0000 $\pm$ 0.0000
& 0.0000 $\pm$ 0.0000
& 0.0000 $\pm$ 0.0000
& 1.0000 $\pm$ 0.0000
& 0.0000 $\pm$ 0.0000
\\

Random-target constraints
& $-0.0136 \pm 0.0095$
& 0.4908 $\pm$ 0.0351
& 0.1492 $\pm$ 0.0520
& \textbf{0.7450 $\pm$ 0.0326}
& 0.1931 $\pm$ 0.1002
& 0.3333 $\pm$ 0.1667
& 0.6678 $\pm$ 0.0689
\\

Opposite-direction constraints
& $-0.0301 \pm 0.0175$
& 0.3907 $\pm$ 0.0748
& 0.1392 $\pm$ 0.0572
& 0.8710 $\pm$ 0.1257
& 0.0859 $\pm$ 0.0971
& \underline{0.6000 $\pm$ 0.1900}
& 0.4906 $\pm$ 0.1443
\\

Direct evidence-to-loss
& \textbf{0.0405 $\pm$ 0.0387}
& \textbf{0.5769 $\pm$ 0.0340}
& 0.1841 $\pm$ 0.2263
& 0.8880 $\pm$ 0.1365
& \underline{0.2419 $\pm$ 0.2281}
& 0.5667 $\pm$ 0.0913
& \underline{0.6724 $\pm$ 0.1461}
\\

One-shot typed revision
& 0.0022 $\pm$ 0.0086
& 0.4562 $\pm$ 0.0639
& \underline{0.0968 $\pm$ 0.0357}
& \underline{0.8160 $\pm$ 0.1394}
& 0.1449 $\pm$ 0.1373
& 0.5667 $\pm$ 0.0913
& 0.5428 $\pm$ 0.0745
\\

All eligible revisions
& 0.0088 $\pm$ 0.0284
& 0.5250 $\pm$ 0.0794
& 0.1414 $\pm$ 0.1149
& 0.8770 $\pm$ 0.0596
& 0.2234 $\pm$ 0.1540
& 0.5333 $\pm$ 0.1394
& 0.6188 $\pm$ 0.1047
\\

\method{}
& \underline{0.0159 $\pm$ 0.0059}
& \underline{0.5568 $\pm$ 0.0551}
& \textbf{0.0859 $\pm$ 0.0370}
& 0.8220 $\pm$ 0.1095
& \textbf{0.2505 $\pm$ 0.0860}
& \textbf{0.6667 $\pm$ 0.1667}
& \textbf{0.6994 $\pm$ 0.0646}
\\

\bottomrule
\end{tabular}
}
\end{table*}

Table~\ref{tab:target_results} shows that direct evidence-to-loss produces the
largest raw target movement, but also the greatest and most variable
off-target change. In contrast, \method{} achieves the lowest non-trivial NOTC
and the highest TRS, PPR, and DPR, while retaining the second-highest NTGC and
TIP. Thus, the complete loop does not maximize target displacement alone; it
provides the strongest overall balance between target improvement, update
specificity, predictive preservation, and validation-to-test persistence.

Random-target and opposite-direction controls yield negative NTGC, indicating
that the benefits cannot be attributed to arbitrary additional regularization.
Compared with one-shot revision, the complete loop obtains higher target
closure, specificity, predictive preservation, and persistence, supporting the
use of maintenance constraints across rounds. Applying all eligible revisions
without response probing also increases off-target change, indicating that
typed probing and portfolio selection help control interference among
candidate revisions.

Detailed strategy definitions, metric construction, and aggregation procedures
are provided in Appendix~\ref{app:rq4_targeted_effect}.

\subsection{RQ5: Iterative Knowledge-guided Refinement}
\label{sec:iterative_results}
We examine how the constraint portfolio evolves as each updated predictor is
re-explained and newly discovered relations undergo evidence verification.
Loops~1--6 constitute the primary refinement trajectory used in the matched
quantitative evaluation in RQ4. Loop~1 introduces a signed-response correction
for bilateral hippocampal connectivity together with an age-aware
latent-geometry constraint. Loop~2 adds longitudinal prediction ordering,
prediction calibration, and interval-ordered latent displacement.
Loops~3--4 address age-specific prediction bias and interval-aware
longitudinal robustness. Loops~5--6 introduce within-subject representation
stability, verified FC importance, and a retuned interval-aware constraint.
After the six-round trajectory, we continued the refinement process for three
additional exploratory rounds. Loops~7--9 introduced further FC bundles and
additional temporal-representation constraints, but did not improve the
validation results beyond the Loop-6 checkpoint. We therefore use Loop~6 as
the validation-selected final predictor in the primary quantitative
comparison, while reporting the later rounds qualitatively to illustrate how
the explanation--knowledge loop continues to generate and evaluate candidate
revisions. Previously accepted revisions are retained as maintenance
constraints throughout the trajectory.
\paragraph{Case study: refinement of an aging-related FC family.}
A later refinement loop identifies low explanatory importance for functional
connections involving the posterior cingulate cortex and precuneus. Literature
verification supports the relevance of these connections to normal aging, but
does not establish a sufficiently reliable positive or negative relation.
\method{} therefore revises their relative attribution importance while
leaving their response directions unconstrained.

Let $\Phi_j$ denote the inherited explanation operator measuring the relative
importance of the verified FC family. Following the general refinement
procedure, the current model relation is minimally projected onto its
knowledge-consistent set:
\begin{equation}
    r_{t,j}^{M}
    =
    \Phi_j
    \left(
        f_{\theta_t}
    \right),
    \qquad
    r_{t,j}^{R}
    =
    \operatorname{Proj}_{\mathcal{C}_{K,t,j}}
    \left(
        r_{t,j}^{M}
    \right).
    \label{eq:rq5_importance_projection}
\end{equation}

The projected relation is realized by penalizing its discrepancy from the
corresponding relation in the updated predictor:
\begin{equation}
    \delta_{t,j}(\theta)
    =
    D_j
    \left(
        \Phi_j(f_{\theta}),
        r_{t,j}^{R}
    \right).
    \label{eq:rq5_importance_realization}
\end{equation}
Here, $\Phi_j$ operates on attribution importance rather than signed response.
The revision therefore increases the model's relative dependence on the
aging-relevant FC family without imposing a direction that is not supported by
the evidence.

This case illustrates how \method{} matches the candidate explanatory revision
to the specificity of the available evidence: verified aging relevance can
support a direction-neutral importance revision, but not an unsupported signed
FC--age relation.

\begin{table*}[!t]
\centering
\caption{Controlled stress tests of the \method{} pipeline. The full row
reports the final selected model, whereas the remaining rows represent
computational counterfactuals designed to expose specific component failure
modes.}
\label{tab:ablation_results}

\scriptsize
\setlength{\tabcolsep}{7pt}
\renewcommand{\arraystretch}{1.05}

\begin{tabular}{@{}lccccc@{}}
\toprule
Variant
& MAE $\downarrow$
& PCC $\uparrow$
& Confirm. $\uparrow$
& Progress $\uparrow$
& Drift $\downarrow$
\\
\midrule

Full \method{}
& \textbf{4.3813 $\pm$ 0.0258}
& \textbf{0.5274 $\pm$ 0.0131}
& \textbf{0.6500 $\pm$ 0.0601}
& \textbf{0.8444 $\pm$ 0.0609}
& \textbf{0.0286 $\pm$ 0.0639}
\\

w/o reliability filtering
& 4.5697 $\pm$ 0.2259
& 0.4923 $\pm$ 0.0412
& 0.4437 $\pm$ 0.0464
& 0.4000 $\pm$ 0.2236
& 0.0571 $\pm$ 0.0782
\\

w/o model-blind questions
& 4.6765 $\pm$ 0.2336
& 0.4747 $\pm$ 0.0260
& 0.4467 $\pm$ 0.0691
& 0.3000 $\pm$ 0.4472
& 0.0286 $\pm$ 0.0639
\\

w/o scope-aware verification
& 4.5885 $\pm$ 0.1366
& 0.4766 $\pm$ 0.0393
& 0.5938 $\pm$ 0.0442
& 0.3000 $\pm$ 0.2739
& 0.0571 $\pm$ 0.0782
\\

w/o minimum projection
& 4.5965 $\pm$ 0.0967
& 0.4637 $\pm$ 0.0484
& 0.4313 $\pm$ 0.1645
& 0.5000 $\pm$ 0.3536
& 0.0286 $\pm$ 0.0639
\\

w/o variable--operator preservation
& 4.6098 $\pm$ 0.1550
& 0.4724 $\pm$ 0.0259
& 0.4581 $\pm$ 0.1503
& 0.3000 $\pm$ 0.4472
& 0.0286 $\pm$ 0.0639
\\

w/o response probing
& 4.5976 $\pm$ 0.2318
& 0.4856 $\pm$ 0.0252
& 0.6125 $\pm$ 0.2292
& 0.8000 $\pm$ 0.2739
& 0.0857 $\pm$ 0.1278
\\

w/o sparse portfolio selection
& 4.4672 $\pm$ 0.0597
& 0.4984 $\pm$ 0.0170
& 0.5625 $\pm$ 0.1398
& 0.8000 $\pm$ 0.2739
& 0.0571 $\pm$ 0.0782
\\

\bottomrule
\end{tabular}
\end{table*}

\subsection{RQ6: Component Ablation}
\label{sec:ablation_results}

We conduct seven component ablations, each removing or modifying
one component of the complete refinement pipeline while retaining the same
data partitions, random seeds, initialization protocol, optimization budget,
and evaluation procedure. Reliability filtering determines which model claims
may enter verification; model-blind questions reduce confirmation bias;
scope-aware verification prevents broad evidence from authorizing overly
specific revisions; minimum projection limits unnecessary intervention; and
variable--operator preservation maintains the semantic correspondence between
verified knowledge and its executable loss. Response probing tests whether a
candidate loss produces the intended model response, whereas sparse portfolio
selection limits interference among otherwise eligible revisions.

The complete system achieves the best value on every commonly reported
metric. Removing reliability filtering substantially reduces both explanation
confirmation and target progress. Exposing the model conclusion during
question generation produces the largest predictive degradation, whereas
ignoring evidence scope preserves relatively high confirmation but yields low
progress, showing that a reproducible model claim is not necessarily supported
at the resolution required for intervention.

Removing response probing or sparse selection retains high raw progress but
introduces detectable off-target change. These results indicate that the
components serve complementary roles: reliability and verification determine
what may be revised, typed projection determines how the revision is defined,
and probing and portfolio selection control whether it can be realized without
unnecessary model change. Operational definitions and interpretation
limitations are provided in Appendix~\ref{app:rq6_ablation}.

\section{Limitations and Ethical Considerations}
\label{sec:limitations_ethics}

Our framework depends on the reliability of post-hoc explanations, literature
retrieval, and evidence verification. Incomplete, conflicting, or mis-scoped
evidence may therefore lead to inappropriate abstention or revision, and
knowledge consistency should not be interpreted as causal or clinical
validation. The present study is limited to resting-state functional
connectivity and requires external evaluation across populations, acquisition
sites, and imaging protocols. Brain-age predictions and explanations may also
reflect demographic or dataset biases and are not intended for individual
clinical decision-making. All analyses use de-identified research data, and
the framework avoids exposing subject-level data to the literature-verification
agent.

\section{Generative AI Usage}
\label{sec:generative_ai_usage}

Generative AI tools, including ChatGPT and coding assistants, were used during
the research process for manuscript editing, organization, code debugging, and
implementation support. They were not treated as sources of scientific
evidence or used to make final scientific decisions. All generated text, code,
literature claims, citations, analyses, and conclusions were reviewed and
verified by the authors, who take full responsibility for the content of this
work.

\section{Conclusion}
\label{sec:conclusion}
We introduced \method{}, a closed loop that turns reliable post-hoc model
behavior into literature-verified, minimally revised explanations and then
into validated training constraints. The central design principle is to keep
scientific evidence, explanation geometry, and optimization distinct but
connected through shared variables, operators, and scopes. The brain-age
instantiation evaluates not only predictive performance but also explanation
reliability, held-out domain alignment, target-directed movement, and
unintended explanatory drift. More broadly, the framework provides a general
mechanism for converting verified scientific knowledge into model-specific
and auditable updates without prespecifying the correction target. This
formulation provides a testable route from model interpretation to automated
scientific model refinement.
\bibliographystyle{ACM-Reference-Format}
\bibliography{xai_refine_compact}

\clearpage
\appendix

\section*{Appendix}

\section{Details of Reliable Post-hoc Explanation Discovery}
\label{app:module1_details}

\subsection{Multi-seed Post-hoc Analysis Protocol}
\label{app:posthoc_protocol}

At iteration $t$, the current predictor is trained under $S$ random seeds,
producing the seed-indexed model family
\begin{equation}
    \mathcal{F}_t
    =
    \left\{
        f_{\theta_t}^{(s)}
    \right\}_{s=1}^{S}.
    \label{eq:app_model_family}
\end{equation}
The same post-hoc analysis and evaluation protocol is applied to every model.
Explanation discovery uses only the data assigned to model development and
does not access the test partition.

For analysis $j\in\mathcal{J}$, let $v_j$ denote the model quantity being
examined, $\Phi_j$ the corresponding measurement operator, and $\Omega_j$ the
scope over which the measurement is evaluated. The seed-wise outputs are
\begin{equation}
    \mathcal{O}_{t,j}
    =
    \left\{
        o_{t,j}^{(s)}
    \right\}_{s=1}^{S},
    \qquad
    o_{t,j}^{(s)}
    =
    \Phi_j
    \left(
        f_{\theta_t}^{(s)};
        v_j,
        \Omega_j
    \right).
    \label{eq:app_seed_outputs}
\end{equation}
Each output belongs to the typed explanation space
$o_{t,j}^{(s)}\in\mathcal{Z}_j$. We use
$\mathcal{O}_{t,j}$ for the collection of seed-wise measurements and reserve
$\mathcal{Z}_j$ for the mathematical space in which the corresponding
explanatory relation is represented.

When multiple scans are available for the same subject, scan-level
measurements are first aggregated within subject and then across subjects.
For a scan-level quantity $q_{i,s}$, the subject-balanced estimate is
\begin{equation}
    \overline{q}_{s}
    =
    \frac{1}{|\mathcal{U}|}
    \sum_{u\in\mathcal{U}}
    \frac{1}{N_u}
    \sum_{i:m_i=u}
    q_{i,s},
    \label{eq:app_subject_balanced}
\end{equation}
where $\mathcal{U}$ is the set of subjects and $N_u$ is the number of scans
from subject $u$. This prevents subjects with more repeated observations from
receiving greater total weight.

\subsection{Post-hoc Analysis Families}
\label{app:analysis_families}

The framework uses complementary post-hoc analyses rather than treating one
output as a complete explanation of the predictor.

\paragraph{Prediction behavior.}
Prediction-level analyses measure overall error, prediction-range compression,
residual dependence on chronological age, age-specific bias, and calibration.
These analyses characterize how the predictor behaves across the observed age
range rather than relying only on an aggregate error metric.

\paragraph{Longitudinal consistency.}
For two temporally ordered scans $i$ and $i'$ from the same subject, define
\begin{equation}
    \Delta y_{ii'}
    =
    y_{i'}-y_i,
    \qquad
    \Delta\widehat{y}_{ii'}^{(s)}
    =
    f_{\theta_t}^{(s)}(\mathbf{x}_{i'})
    -
    f_{\theta_t}^{(s)}(\mathbf{x}_{i}).
    \label{eq:app_longitudinal_change}
\end{equation}
The corresponding operators examine the direction and magnitude of predicted
within-subject aging, residual consistency across repeated scans, and the
dependence of longitudinal error on follow-up duration.

\paragraph{Feature response.}
Feature-level analyses characterize how predicted age changes with FC edges,
ROIs, or network-level feature groups. Signed response, absolute response
magnitude, and relative importance are retained as distinct explanatory
quantities because they encode different scientific relations.

\paragraph{Attribution structure.}
Attribution analyses examine the distribution, concentration, sign, and
anatomical organization of the predictor's feature dependence. When several
attribution methods are used, their outputs are retained separately unless
their mathematical quantities and scopes are compatible.

\paragraph{Representation geometry.}
Representation-level analyses examine whether latent distances preserve
age-related organization, repeated-scan similarity, subject identity, and
longitudinal displacement. Depending on the operator, the resulting relation
may be a scalar association, pairwise ordering, distance matrix, or local
neighborhood structure.

\subsection{Cross-seed Aggregation}
\label{app:cross_seed_aggregation}

The model-side relation reported for analysis $j$ is
\begin{equation}
    r_{t,j}^{M}
    =
    \operatorname{Agg}_{j}
    \left(
        \mathcal{O}_{t,j}
    \right)
    \in
    \mathcal{Z}_j.
    \label{eq:app_model_relation}
\end{equation}
The aggregation operator is chosen according to the relation type. Scalar
measurements may use a robust location estimator, vector-valued measurements
may use coordinate-wise robust aggregation, rankings may use rank aggregation,
and structured relations may use an analysis-specific consensus operator.

Aggregation is source-preserving. It may summarize compatible seed-wise
measurements, but it cannot alter the analyzed variable, measurement operator,
scope, direction, numerical values, or provenance of the source outputs. The
complete seed-wise collection $\mathcal{O}_{t,j}$ is retained as supporting
evidence for $r_{t,j}^{M}$.

\subsection{Cross-seed Reproducibility}
\label{app:pattern_evaluation}

Every seed-specific output is assigned one of four evaluation states relative
to the candidate relation $r_{t,j}^{M}$:
\begin{itemize}
    \item \emph{supporting}, when it supports the candidate relation;
    \item \emph{contradicting}, when it supports an incompatible relation;
    \item \emph{inconclusive}, when neither conclusion is sufficiently supported;
    \item \emph{unavailable}, when the required measurement cannot be evaluated.
\end{itemize}

Supporting, contradicting, and inconclusive outputs are treated as evaluable.
Let $N_{t,j}^{\mathrm{sup}}$ be the number of supporting seeds and
$N_{t,j}^{\mathrm{eval}}$ the number of evaluable seeds. Support and coverage
are
\begin{equation}
    \rho_{t,j}^{\mathrm{sup}}
    =
    \frac{
        N_{t,j}^{\mathrm{sup}}
    }{
        N_{t,j}^{\mathrm{eval}}
    },
    \qquad
    \rho_{t,j}^{\mathrm{cov}}
    =
    \frac{
        N_{t,j}^{\mathrm{eval}}
    }{
        S
    }.
    \label{eq:app_support_coverage}
\end{equation}

For scalar measurements, numerical variation is measured using the median
absolute deviation
\begin{equation}
    \operatorname{MAD}_{t,j}
    =
    \operatorname{median}_{s}
    \left|
        o_{t,j}^{(s)}
        -
        \operatorname{median}_{s'}
        o_{t,j}^{(s')}
    \right|.
    \label{eq:app_mad}
\end{equation}
A normalized scalar stability score is
\begin{equation}
    \rho_{t,j}^{\mathrm{stab}}
    =
    1
    -
    \frac{
        \operatorname{MAD}_{t,j}
    }{
        \operatorname{median}_{s}
        \left|
            o_{t,j}^{(s)}
        \right|
        +
        \operatorname{MAD}_{t,j}
        +
        \epsilon
    }.
    \label{eq:app_stability}
\end{equation}
For vector, ranking, distributional, or geometric outputs, this scalar score is
replaced by a type-compatible agreement measure.

The overall reproducibility follows the weakest-link principle:
\begin{equation}
    \rho_{t,j}
    =
    \min
    \left\{
        \rho_{t,j}^{\mathrm{sup}},
        \rho_{t,j}^{\mathrm{cov}},
        \rho_{t,j}^{\mathrm{stab}}
    \right\}.
    \label{eq:app_reproducibility}
\end{equation}
Thus, a high score requires the relation to be supported among evaluable runs,
broadly measurable across runs, and stable in its corresponding explanation
space. Directional relations additionally require cross-seed sign agreement.
Effect magnitude is retained in $r_{t,j}^{M}$ and is not folded into
$\rho_{t,j}$.

\subsection{Faithfulness Validation}
\label{app:faithfulness_validation}

Cross-seed reproducibility does not establish that a post-hoc relation
accurately characterizes the predictor. Each analysis therefore has an
associated validation operator
\begin{equation}
    \operatorname{Fid}_{t,j}
    =
    \operatorname{Validate}_{j}
    \left(
        r_{t,j}^{M},
        f_{\theta_t};
        v_j,
        \Omega_j
    \right)
    \in
    [0,1].
    \label{eq:app_faithfulness}
\end{equation}

The validation procedure is selected according to the explanatory quantity.
Feature-response relations may be validated by controlled input perturbation;
attribution relations may use model- and input-randomization tests;
prediction-level relations may be compared directly with observed model
outputs; longitudinal relations may be recomputed from held-out within-subject
pairs; and representation relations may be remeasured from independently
trained models.

The validation operator must evaluate the same relation represented by
$\Phi_j$. A high faithfulness score therefore indicates that the explanation
agrees with the actual behavior of the predictor, rather than merely appearing
stable across model runs.

\subsection{Model-valid Explanation Records}
\label{app:structured_explanations}

An analysis enters the verification loop only when it satisfies both the
reproducibility and faithfulness requirements:
\begin{equation}
    \mathcal{J}_{t}^{M}
    =
    \left\{
        j\in\mathcal{J}
        \,\middle|\,
        \rho_{t,j}\geq\tau_{\rho},
        \;
        \operatorname{Fid}_{t,j}\geq\tau_F
    \right\}.
    \label{eq:app_retained_analyses}
\end{equation}

For every retained analysis, the model-side explanation record is
\begin{equation}
    e_{t,j}^{M}
    =
    \left(
        v_j,
        \Phi_j,
        \Omega_j,
        r_{t,j}^{M},
        \rho_{t,j},
        \operatorname{Fid}_{t,j}
    \right),
    \qquad
    j\in\mathcal{J}_{t}^{M}.
    \label{eq:app_model_explanation_record}
\end{equation}
The complete model-side explanation state is
\begin{equation}
    z_t^{M}
    =
    \left\{
        e_{t,j}^{M}
        \,\middle|\,
        j\in\mathcal{J}_{t}^{M}
    \right\}.
    \label{eq:app_model_explanation_state}
\end{equation}

Records that fail either threshold remain in the audit trail but do not produce
literature queries or model-refinement constraints. Natural-language
statements are generated only as readable views of these typed records and are
not used as the persistent internal representation.

\subsection{Source-preserving Consolidation}
\label{app:explanation_consolidation}

Several atomic records may describe compatible aspects of a higher-level
scientific pattern. Such records may be grouped for presentation and
literature-search efficiency, but every executable record remains linked to its
atomic source analysis.

Let $\mathcal{G}_{t,k}$ denote a compatible group of source records. A
consolidated readable claim may summarize $\mathcal{G}_{t,k}$ only when its
members agree in scientific entity, relation type, scope, and direction. Its
reliability cannot exceed that of its least reliable source:
\begin{equation}
    \rho_{t,k}^{\mathrm{cons}}
    =
    \min_{
        e_{t,j}^{M}\in\mathcal{G}_{t,k}
    }
    \rho_{t,j}.
    \label{eq:app_consolidated_reliability}
\end{equation}
Consolidation cannot create a new model variable, change a measurement
operator, broaden an anatomical or population scope, alter an effect
direction, or replace source measurements. Literature evidence retrieved for a
shared question is returned to each atomic explanation record for separate
scope alignment and knowledge-set compilation.

\section{Details of Literature Verification and Knowledge-Set Compilation}
\label{app:module2_details}

\subsection{Neutral Scientific-Question Construction}
\label{app:neutral_question_generation}

For each retained explanation $e_{t,j}^{M}\in z_t^{M}$, the query operator
constructs
\begin{equation}
    q_{t,j}
    =
    \mathcal{Q}
    \left(
        e_{t,j}^{M}
    \right).
    \label{eq:app_question_generation}
\end{equation}
Operationally, $\mathcal{Q}$ uses the scientific variable $v_j$, the relation
type represented by $\Phi_j$, and the applicable scope $\Omega_j$. The
model-observed relation $r_{t,j}^{M}$, its effect magnitude, reproducibility
score, and faithfulness score are withheld whenever revealing them could bias
retrieval toward confirmation.

For example, if the model exhibits a positive age-related response for a
hippocampal FC pattern, the generated question asks how the corresponding FC
pattern varies with age in a comparable population. It does not state that the
current predictor observed a positive relation.

Identical neutral questions may share a literature search, but every retrieved
record is returned to each originating explanation for separate entity, scope,
and applicability assessment.

\subsection{Model-blind Literature Retrieval}
\label{app:model_blind_retrieval}

The retrieval stage receives the neutral question and the scientific scope
needed to identify relevant studies. It does not receive model parameters,
predictions, subject-level data, the model-observed direction, or downstream
training objectives.

For query $q_{t,j}$, let
\begin{equation}
    \mathcal{R}_{t,j}^{\mathrm{lit}}
    =
    \operatorname{Retrieve}
    \left(
        q_{t,j};
        \mathcal{L}_{\mathrm{lit}}
    \right)
    \label{eq:app_literature_retrieval}
\end{equation}
denote the retrieved candidate records. Each candidate stores its
bibliographic identifier, title, reported population, imaging modality,
anatomical scope, scientific construct, reported relation, and relevant
evidence passage.

Retrieval recall and evidence verification are treated as separate operations.
A retrieved document does not contribute to the knowledge record until its
bibliographic and scientific content have been verified.

\subsection{Evidence Verification}
\label{app:evidence_verification}

The verification operator checks each retrieved source for:

\begin{itemize}
    \item bibliographic identity and availability;
    \item correspondence between the cited passage and the source;
    \item scientific-variable and construct match;
    \item population, modality, and measurement compatibility;
    \item anatomical specificity, including laterality when applicable;
    \item the direction, magnitude, uncertainty, or null status of the reported
    relation.
\end{itemize}

Evidence is assigned only to the most specific scope justified by the source.
For example, evidence concerning a broad anatomical system cannot be treated
as connection-specific evidence for an individual bilateral FC edge.
Similarly, cross-sectional age associations and within-subject longitudinal
aging effects remain distinct scientific relations.

Let $\mathcal{B}_{t,j}$ denote the verified evidence bundle retained for query
$j$. Each element of $\mathcal{B}_{t,j}$ preserves the bibliographic
identifier, evidence passage, extracted relation, supported scope, and
verification outcome.

\subsection{Structured Knowledge Records}
\label{app:structured_knowledge_records}

The verification operator produces
\begin{equation}
\label{eq:app_knowledge_record}
    \kappa_{t,j}
    =
    \left(
        v_j,
        \Omega_{t,j}^{K},
        r_{t,j}^{K},
        c_{t,j},
        s_{t,j},
        \mathcal{B}_{t,j}
    \right).
\end{equation}
Here, $\Omega_{t,j}^{K}$ is the scope supported by the verified evidence,
$r_{t,j}^{K}$ is the corresponding knowledge-side relation, $c_{t,j}\in[0,1]$
is the calibrated evidence confidence, $s_{t,j}$ is the evidence status, and
$\mathcal{B}_{t,j}$ provides provenance.

The evidence status distinguishes directional evidence, verified null
evidence, mixed evidence, insufficient evidence, and unavailable evidence.
A verified null relation is not equivalent to a failed search. The former may
bound the allowed magnitude of a model relation, whereas the latter produces no
literature-driven restriction.

The complete knowledge state is
\begin{equation}
    \kappa_t
    =
    \left\{
        \kappa_{t,j}
        \,\middle|\,
        e_{t,j}^{M}\in z_t^{M}
    \right\}.
    \label{eq:app_knowledge_state}
\end{equation}

\subsection{Scope Alignment}
\label{app:scope_alignment}

The model-side scope $\Omega_j$ and the literature-supported scope
$\Omega_{t,j}^{K}$ are aligned through
\begin{equation}
    \Omega_{t,j}^{\cap}
    =
    \operatorname{Align}
    \left(
        \Omega_j,
        \Omega_{t,j}^{K}
    \right).
    \label{eq:app_scope_alignment}
\end{equation}
Alignment checks population, age range, imaging modality, anatomical entity,
measurement definition, temporal interpretation, and scientific relation type.

The aligned scope need not equal either original scope. It represents only the
portion for which the model-side and knowledge-side relations are scientifically
comparable. If no scientifically valid overlap exists, then
$\Omega_{t,j}^{\cap}=\varnothing$ and the record is deferred.

Scope alignment never changes $v_j$ or $\Phi_j$. These objects remain inherited
from the originating model explanation so that any downstream revision is
applied to the same model quantity that generated the original hypothesis.

\subsection{Knowledge-set Compilation}
\label{app:knowledge_set_compilation}

A deterministic compiler translates the structured knowledge record into the
same explanation space as the model relation:
\begin{equation}
    \mathcal{C}_{K,t,j}
    =
    \operatorname{Compile}
    \left(
        \kappa_{t,j};
        \mathcal{Z}_j,
        \Omega_{t,j}^{\cap}
    \right)
    \subseteq
    \mathcal{Z}_j.
    \label{eq:app_knowledge_set}
\end{equation}

The compiler uses the verified relation, evidence uncertainty, evidence
status, and aligned scope. Representative cases include:

\paragraph{Directional scalar relation.}
If applicable evidence supports sign $\sigma_{t,j}\in\{-1,+1\}$ with margin
$m_{t,j}\geq0$, then
\begin{equation}
    \mathcal{C}_{K,t,j}
    =
    \left\{
        r\in\mathbb{R}
        :
        \sigma_{t,j}r
        \geq
        m_{t,j}
    \right\}.
    \label{eq:app_directional_set}
\end{equation}

\paragraph{Interval-valued relation.}
If the evidence supports an admissible interval, then
\begin{equation}
    \mathcal{C}_{K,t,j}
    =
    \left[
        \ell_{t,j},
        u_{t,j}
    \right].
    \label{eq:app_interval_set}
\end{equation}

\paragraph{Verified null relation.}
If strong evidence supports the absence of a substantial effect, then
\begin{equation}
    \mathcal{C}_{K,t,j}
    =
    \left\{
        r:
        \|r\|_j
        \leq
        b_{t,j}
    \right\},
    \label{eq:app_null_set}
\end{equation}
where $b_{t,j}$ reflects the uncertainty and resolution of the evidence.

\paragraph{Ordering or relational constraint.}
For pairwise or geometric relations, the admissible set may be
\begin{equation}
    \mathcal{C}_{K,t,j}
    =
    \left\{
        r:
        h_{t,j}(r)
        \leq
        0
    \right\},
    \label{eq:app_relational_set}
\end{equation}
where $h_{t,j}$ encodes the verified ordering, margin, invariance, or geometric
condition.

Strong and directly applicable evidence may produce a narrow set. Broader,
mixed, or uncertain evidence produces a wider set. When no applicable evidence
is available, the compiler returns
\begin{equation}
    \mathcal{C}_{K,t,j}
    =
    \mathcal{Z}_j,
    \label{eq:app_no_update_set}
\end{equation}
which induces no literature-driven revision.

For an executable revision, the compiler must return a nonempty closed set
under the selected discrepancy $D_j$. If the evidence cannot be represented by
such a set, or if the aligned scope is empty, the record is deferred rather
than converted into a training constraint.

\subsection{Guidance Confidence and Eligibility}
\label{app:guidance_confidence}

The guidance confidence is
\begin{equation}
    w_{t,j}
    =
    G
    \left(
        \rho_{t,j},
        \operatorname{Fid}_{t,j},
        c_{t,j},
        a_{t,j}
    \right)
    \in
    [0,1],
    \label{eq:app_guidance_confidence}
\end{equation}
where $a_{t,j}$ summarizes the applicability of the aligned scope and
$G$ is a calibrated monotone aggregation function.

The confidence cannot exceed what is justified by the weakest critical
component. In particular, a highly reproducible model explanation does not
compensate for inapplicable literature, and strong literature evidence does not
compensate for an unfaithful model explanation.

The guidance record passed to projection is
\begin{equation}
    \gamma_{t,j}
    =
    \left(
        v_j,
        \Phi_j,
        \Omega_{t,j}^{\cap},
        \mathcal{C}_{K,t,j},
        w_{t,j}
    \right).
    \label{eq:app_guidance_record}
\end{equation}
The confidence $w_{t,j}$ determines eligibility and controls the allowable
optimization budget after calibration. It is not a scientific target and is
not used as an uncalibrated raw loss coefficient.

\subsection{Abstention and Audit Trail}
\label{app:verification_abstention}

A record produces no executable revision when:

\begin{itemize}
    \item the model-side explanation fails reliability or faithfulness checks;
    \item no applicable literature evidence is found;
    \item the model-side and evidence-side scopes cannot be aligned;
    \item the evidence is too mixed or uncertain to define an executable set;
    \item the compiled admissible set is empty or not closed under the selected
    projection geometry;
    \item the resulting guidance confidence is below $\tau_w$.
\end{itemize}

Abstention does not delete the record. The framework retains the original
explanation, neutral question, retrieved candidates, verified evidence,
knowledge record, alignment decision, compiled set, and abstention reason.
This preserves a complete path from model diagnosis to downstream revision or
deferral.

\section{Details of Minimal Revision and Model Realization}
\label{app:module3_details}

\subsection{Typed Minimum-change Projection}
\label{app:minimal_revision}

For each executable guidance record, the revised relation is
\begin{equation}
    r_{t,j}^{R}
    =
    \operatorname{Proj}_{\mathcal{C}_{K,t,j}}
    \left(
        r_{t,j}^{M}
    \right)
    \in
    \arg\min_{
        r\in\mathcal{C}_{K,t,j}
    }
    D_j
    \left(
        r,
        r_{t,j}^{M}
    \right).
    \label{eq:app_projection}
\end{equation}

The verified evidence determines the admissible set, while the current model
relation supplies the reference state. The projection therefore changes the
explanation only as much as required for knowledge consistency.

When
$r_{t,j}^{M}\in\mathcal{C}_{K,t,j}$, the projection leaves it unchanged:
\begin{equation}
    r_{t,j}^{R}
    =
    r_{t,j}^{M}.
    \label{eq:app_projection_identity}
\end{equation}
The literature does not directly provide the numerical target
$r_{t,j}^{R}$. It defines the admissible region from which the nearest
model-specific revision is selected.

For Euclidean explanations and a nonempty closed admissible set, a projection
exists. If the set is additionally convex and $D_j$ is strictly convex, the
projection is unique. For non-Euclidean or discrete relations, the framework
uses a type-compatible projection procedure and defers records for which a
well-defined minimum cannot be obtained.

\subsection{Type-specific Discrepancy Functions}
\label{app:typed_discrepancy}

The discrepancy $D_j$ is selected according to the geometry of
$\mathcal{Z}_j$.

\paragraph{Scalar and vector relations.}
For a scalar or vector explanatory relation, a compatible discrepancy is
\begin{equation}
    D_j(a,b)
    =
    \frac{
        \|a-b\|_{p}
    }{
        s_j+\epsilon
    },
    \label{eq:app_vector_discrepancy}
\end{equation}
where $s_j$ is an analysis-specific scale estimated from seed-wise or
validation variability.

\paragraph{Ranking relations.}
For ordered feature or region sets, $D_j$ may measure weighted pairwise
disagreement, rank correlation loss, or top-$k$ set disagreement.

\paragraph{Distributional relations.}
For attribution or response distributions, compatible discrepancies include
Wasserstein distance, maximum mean discrepancy, or another divergence whose
domain matches the representation.

\paragraph{Geometric relations.}
For distance matrices or representation geometry, $D_j$ may measure normalized
matrix distortion, pairwise-order violation, or disagreement between local
neighborhood structures.

The same $D_j$ used to define the minimum revision is also used, whenever
possible, to evaluate whether the updated predictor realizes that revision.

\subsection{Revised Explanation State}
\label{app:revised_explanation_state}

The revised explanation record is
\begin{equation}
    e_{t,j}^{R}
    =
    \left(
        v_j,
        \Phi_j,
        \Omega_{t,j}^{\cap},
        r_{t,j}^{R},
        w_{t,j}
    \right),
    \label{eq:app_revised_record}
\end{equation}
and the complete revised explanation state is
\begin{equation}
    z_t^{R}
    =
    \left\{
        e_{t,j}^{R}
        \,\middle|\,
        e_{t,j}^{M}\in z_t^{M}
    \right\}.
    \label{eq:app_revised_state}
\end{equation}

The model variable and measurement operator are inherited without
reinterpretation:
\begin{equation}
    v_j^{\mathrm{discovery}}
    \equiv
    v_j^{\mathrm{refinement}},
    \qquad
    \Phi_j^{\mathrm{discovery}}
    \equiv
    \Phi_j^{\mathrm{refinement}}.
    \label{eq:app_operator_inheritance}
\end{equation}
Consequently, a discrepancy diagnosed in predictions, feature responses,
longitudinal behavior, attributions, or latent geometry is corrected on the
same quantity and evaluated in the same explanatory dimension.

\subsection{Actionable Revision Filtering}
\label{app:revision_eligibility}

The projection gap is
\begin{equation}
    g_{t,j}^{\mathrm{proj}}
    =
    D_j
    \left(
        r_{t,j}^{M},
        r_{t,j}^{R}
    \right).
    \label{eq:app_projection_gap}
\end{equation}
A revision is actionable only when the projection is non-negligible,
sufficiently reliable, and associated with a valid aligned scope:
\begin{equation}
    \mathcal{A}_t
    =
    \left\{
        j
        \,\middle|\,
        g_{t,j}^{\mathrm{proj}}
        >
        \varepsilon_{\mathrm{eq}},
        \;
        w_{t,j}\geq\tau_w,
        \;
        \Omega_{t,j}^{\cap}\neq\varnothing
    \right\}.
    \label{eq:app_actionable_set}
\end{equation}

An index excluded from $\mathcal{A}_t$ remains in the audit trail. Exclusion may
mean that the model already satisfies the verified relation, that the
projection is smaller than the practical resolution threshold, or that the
guidance is not sufficiently reliable or applicable.

\subsection{Revision Realization}
\label{app:revision_realization}

For each $j\in\mathcal{A}_t$, the candidate model is remeasured using the
inherited operator:
\begin{equation}
    \widehat{r}_{t,j}(\theta)
    =
    \Phi_j
    \left(
        f_{\theta};
        v_j,
        \Omega_{t,j}^{\cap}
    \right).
    \label{eq:app_candidate_relation}
\end{equation}
Its realization discrepancy is
\begin{equation}
    \delta_{t,j}(\theta)
    =
    D_j
    \left(
        \widehat{r}_{t,j}(\theta),
        r_{t,j}^{R}
    \right).
    \label{eq:app_realization_discrepancy}
\end{equation}

The conceptual model update is
\begin{equation}
\label{eq:app_constrained_update}
\begin{aligned}
    \widetilde{\theta}_{t+1}
    \in
    \arg\min_{\theta}\quad
    &
    \mathcal{L}_{\mathrm{pred}}
    \left(
        \theta;
        \mathcal{D}_{\mathrm{tr}}
    \right)
    \\
    \mathrm{s.t.}\quad
    &
    \delta_{t,j}(\theta)
    \leq
    \varepsilon_{R,j},
    \qquad
    j\in\mathcal{A}_t.
\end{aligned}
\end{equation}

Thus, prediction remains the optimized objective, while realization of the
revised explanations is treated as a collection of admissibility constraints.

\subsection{Differentiable Surrogate Construction}
\label{app:surrogate_realization}

Some operators $\Phi_j$ are directly differentiable, whereas others involve
discrete rankings, perturbation experiments, repeated model fitting, or
non-differentiable aggregation. For a non-differentiable analysis, training
uses a surrogate
\begin{equation}
    \widetilde{\Phi}_j
    \approx
    \Phi_j,
    \qquad
    \widetilde{\delta}_{t,j}(\theta)
    =
    \widetilde{D}_j
    \left(
        \widetilde{\Phi}_j
        \left(
            f_{\theta};
            v_j,
            \Omega_{t,j}^{\cap}
        \right),
        r_{t,j}^{R}
    \right).
    \label{eq:app_surrogate_discrepancy}
\end{equation}

A valid surrogate must satisfy two requirements. First, reducing
$\widetilde{\delta}_{t,j}$ should induce movement in the intended direction
under the original operator. Second, the original
$\Phi_j$ and $D_j$ must be reapplied after training to determine whether the
revision was actually realized.

Representative realizations include signed hinge losses for directional
responses, robust target matching for scalar or vector relations, pairwise
ranking losses for ordering relations, contrastive or triplet objectives for
representation geometry, and distributional alignment for attribution
patterns.

\subsection{Confidence-to-budget Calibration}
\label{app:guidance_scaling}

The scientific guidance confidence $w_{t,j}$ is not used directly as an
optimization coefficient. A calibrated mapping
\begin{equation}
    \alpha_{t,j}
    =
    \alpha
    \left(
        w_{t,j};
        j
    \right)
    \label{eq:app_confidence_mapping}
\end{equation}
converts confidence into an analysis-specific training budget.

The mapping accounts for both scientific confidence and the numerical scale of
the selected violation measure. Therefore, two revisions with equal
$w_{t,j}$ need not receive identical raw coefficients when their unscaled
gradients or discrepancy ranges differ substantially.

The differentiable implementation is
\begin{equation}
\label{eq:app_penalty_update}
    \widetilde{\theta}_{t+1}
    \in
    \arg\min_{\theta}
    \left[
        \mathcal{L}_{\mathrm{pred}}
        \left(
            \theta;
            \mathcal{D}_{\mathrm{tr}}
        \right)
        +
        \lambda
        \sum_{j\in\mathcal{A}_t}
        \alpha_{t,j}
        \left[
            \widetilde{\delta}_{t,j}(\theta)
            -
            \varepsilon_{R,j}
        \right]_{+}
    \right],
\end{equation}
where $[a]_{+}=\max(0,a)$.

\subsection{Candidate Response Probing}
\label{app:response_probing}

A mathematically compatible surrogate may still fail to move the current
predictor in the intended direction. Before full retraining, the framework
therefore performs a short response probe for each candidate surrogate and
budget.

The probe measures:

\begin{itemize}
    \item movement of the original explanatory relation toward
    $r_{t,j}^{R}$;
    \item reduction relative to the initial projection gap;
    \item change in validation prediction performance;
    \item gradient finiteness and model-state validity;
    \item off-target change in non-revised explanatory quantities.
\end{itemize}

Each probe is compared with a sham continuation using the same parent
checkpoint and optimization schedule but without the revision constraint.
This distinguishes constraint-induced movement from ordinary continuation of
training.

If several surrogates are compatible with the same revision, the framework
retains the surrogate and budget that produce the strongest valid movement
under the original operator while satisfying the predictive guardrail.

\subsection{Sparse Revision Portfolio}
\label{app:portfolio_construction}

Individually actionable revisions are not necessarily jointly compatible.
The implementation may therefore select a sparse portfolio
\begin{equation}
    \mathcal{P}_t
    \subseteq
    \mathcal{A}_t
    \label{eq:app_revision_portfolio}
\end{equation}
based on response-probe success, prediction cost, scientific confidence,
operator overlap, target compatibility, and gradient interaction.

When no portfolio selection is used,
$\mathcal{P}_t=\mathcal{A}_t$. When several explanation records induce the same
variable, operator, scope, revised relation, and surrogate, they may share one
executable term while retaining separate scientific provenance.

With portfolio selection, Equation~\eqref{eq:app_penalty_update} is evaluated
over $j\in\mathcal{P}_t$.

\subsection{Cross-seed Promotion}
\label{app:promotion}

The selected revision portfolio is retrained from the corresponding
seed-specific parent models. For each seed, the framework evaluates prediction
performance and remeasures every targeted relation using its original
$\Phi_j$.

A candidate update must satisfy three validation conditions:
\begin{equation}
    \Delta_{\mathrm{pred}}
    \leq
    \varepsilon_{\mathrm{pred}},
    \qquad
    \Delta_{\mathrm{target}}
    \leq
    -\varepsilon_{\mathrm{target}},
    \qquad
    \Delta_{\mathrm{off}}
    \leq
    \varepsilon_{\mathrm{off}},
    \label{eq:app_promotion_conditions}
\end{equation}
where $\Delta_{\mathrm{pred}}$ is validation prediction degradation,
$\Delta_{\mathrm{target}}$ is the change in targeted realization discrepancy,
and $\Delta_{\mathrm{off}}$ measures non-target explanatory drift.

The intended target movement must also be reproducible across the required
number of validation seeds. A candidate that improves prediction without
moving the intended explanations, or moves the explanations with unacceptable
prediction degradation or off-target drift, is rejected.

The promoted model is
\begin{equation}
    f_{\theta_{t+1}}
    =
    \begin{cases}
        f_{\widetilde{\theta}_{t+1}},
        &
        \text{if the promotion conditions are satisfied},
        \\
        f_{\theta_t},
        &
        \text{otherwise}.
    \end{cases}
    \label{eq:app_model_promotion}
\end{equation}

\subsection{Equilibrium and Loop Termination}
\label{app:loop_termination}

Define the eligible and applicable analysis set
\begin{equation}
    \mathcal{J}_{t}^{\mathrm{elig}}
    =
    \left\{
        j
        \,\middle|\,
        w_{t,j}\geq\tau_w,
        \;
        \Omega_{t,j}^{\cap}\neq\varnothing
    \right\}.
    \label{eq:app_eligible_set}
\end{equation}

An evidence-consistent explanation equilibrium is reached when
\begin{equation}
    r_{t,j}^{M}
    \in
    \mathcal{C}_{K,t,j},
    \qquad
    \forall
    j\in
    \mathcal{J}_{t}^{\mathrm{elig}},
    \label{eq:app_equilibrium_membership}
\end{equation}
or equivalently,
\begin{equation}
    r_{t,j}^{R}
    =
    r_{t,j}^{M},
    \qquad
    \forall
    j\in
    \mathcal{J}_{t}^{\mathrm{elig}}.
    \label{eq:app_equilibrium_projection}
\end{equation}

At equilibrium, every eligible model relation already lies in its applicable
knowledge-consistent set, so the minimum-change projection produces no
knowledge-driven revision.

The condition $\mathcal{A}_t=\varnothing$ is weaker than equilibrium. It may
also occur because evidence is unavailable, confidence is below threshold,
scope alignment fails, the projection is below practical resolution, or no
executable surrogate exists.

The loop terminates when one of the following conditions holds:

\begin{itemize}
    \item evidence-consistent explanation equilibrium is reached;
    \item no actionable and executable revision remains under the current
    evidence and operation library;
    \item no candidate satisfies the promotion conditions for a predefined
    number of consecutive iterations;
    \item progress falls below a predefined tolerance;
    \item the maximum number of iterations is reached.
\end{itemize}

The test set is not used for query generation, response probing, portfolio
selection, promotion, or loop termination. It is evaluated only after the
complete refinement trajectory has been fixed.

\section{Details of Compared Methods}
\label{app:compared_methods}

\paragraph{Predictive baselines.}
We organize the predictive baselines into three groups, with methods within
each group listed approximately by publication year.

\emph{Classical vector-based baselines} include Ridge regression, RBF-SVR,
Random Forest, Elastic Net, and XGBoost. These methods operate on the
vectorized upper triangle of the $81\times81$ FC matrix.

\emph{General neural and graph baselines} include an MLP applied to vectorized
FC features, a standard graph convolutional network (GCN), a graph attention
network (GAT), and the prediction-only TransformerConv backbone $f_0$ used by
our framework.

\emph{Brain-network-specific baselines} include BrainNetCNN,
BrainGNN ~\cite{li2021braingnn},  Brain Network
Transformer ~\cite{kan2022brain}, the 2023 rs-fMRI brain-age
GNN~\cite{gao2023brain}, the 2025 FC Foundation Transformer, and the
Biologically Plausible Brain Graph Transformer (BioBGT). For architectures
originally developed for classification, only the task-specific output head is
replaced by a scalar regression head; the remaining architecture is retained.

All predictive baselines are trained from scratch using the same
subject-disjoint data partitions. Neural models use matched random seeds,
prediction targets, model-selection data, and evaluation procedures whenever
their architectures permit.

\paragraph{Controlled refinement baselines.}
The refinement comparison uses the same prediction-only TransformerConv model
$f_0$ as the common reference. All variants receive matched data splits,
random seeds, optimization settings, early-stopping criteria, and refinement
budgets. We evaluate the following variants:

\begin{itemize}
    \item \textbf{Prediction only}: the reference model $f_0$ without any
    explanation-guided update.

    \item \textbf{Sham revision}: continued training under the same budget but
    without verified guidance or an explanation-revision constraint.

    \item \textbf{Random-target constraints}: constraints with the same number
    and mathematical form as the proposed revisions, but applied to randomly
    selected explanatory targets.

    \item \textbf{Opposite-direction constraints}: the proposed refinement
    procedure with the direction of each revision intentionally reversed.

    \item \textbf{Direct evidence-to-loss}: verified literature evidence is
    translated directly into a training loss without first projecting the
    current model explanation onto a minimally revised state.

    \item \textbf{One-shot typed revision}: explanation guidance is constructed
    once and applied without re-analyzing the updated predictor.

    \item \textbf{All eligible revisions}: all actionable projected revisions are
    applied simultaneously without response probing or actionable-revision
    filtering.

    \item \textbf{\method{}}: the complete iterative procedure, including
    reliable explanation discovery, literature-grounded guidance construction,
    minimal explanation revision, validation-gated model refinement, and
    re-analysis of the updated predictor.
\end{itemize}

These comparisons distinguish improvements caused by verified,
target-specific, and iterative explanation refinement from those attributable
to additional training, random regularization, direct evidence injection, or
the TransformerConv backbone itself.

\section{Detailed Evaluation Metrics and Statistical Protocol}
\label{app:evaluation_metrics}

The evaluation contains eight primary metrics organized into four dimensions:
\begin{equation}
\begin{aligned}
    \text{Prediction}
    &:\
    \operatorname{MAE},
    \operatorname{PCC},
    \\
    \text{Explanation}
    &:\
    \operatorname{Confirmation},
    \operatorname{Faithfulness},
    \\
    \text{Domain alignment}
    &:\
    \operatorname{CSR},
    \operatorname{AEA},
    \\
    \text{Controlled revision}
    &:\
    \operatorname{Progress},
    \operatorname{Drift}.
\end{aligned}
\label{eq:eight_metric_dimensions}
\end{equation}

\subsection{Prediction Metrics}
\label{app:prediction_metrics}

\paragraph{Mean absolute error.}
For $N$ scans with chronological ages $y_i$ and predictions
$\widehat y_i$, mean absolute error is
\begin{equation}
    \operatorname{MAE}
    =
    \frac{1}{N}
    \sum_{i=1}^{N}
    \left|
        \widehat y_i-y_i
    \right|.
    \label{eq:metric_mae}
\end{equation}
MAE is computed at the scan level on the fixed subject-disjoint test split;
repeated scans from the same subject are not averaged before evaluation.
Lower values indicate more accurate age prediction.

\paragraph{Pearson correlation coefficient.}
Let $\overline y$ and $\overline{\widehat y}$ denote the sample means. Pearson
correlation is
\begin{equation}
    \operatorname{PCC}
    =
    \frac{
        \sum_i
        (y_i-\overline y)
        (\widehat y_i-\overline{\widehat y})
    }{
        \sqrt{
            \sum_i
            (y_i-\overline y)^2
        }
        \sqrt{
            \sum_i
            (\widehat y_i-\overline{\widehat y})^2
        }
    }.
    \label{eq:metric_pcc}
\end{equation}
PCC measures whether predicted age follows the linear ordering and trend of
chronological age. It is complementary to MAE: a predictor may achieve high
PCC while retaining a systematic prediction offset. PCC is undefined when
predictions are constant, in which case the number of valid seeds is reported.
All prediction metrics are reported over the same prespecified set of
independently trained runs for every stochastic method. The seed set was fixed
before inspecting the final \method{} results, and no run or seed group was
selected according to validation or test performance. We report the mean and
sample standard deviation across all runs in this fixed set.
\subsection{Explanation Metrics}
\label{app:explanation_metrics}

\subsubsection{Frozen Explanation Claims}

Explanation claims are generated independently for each evaluated model using
the validation split and five discovery seeds
$\mathcal{S}_{\mathrm{disc}}$.

For FC edge $e$ and discovery seed $s$, let
\begin{equation}
    a_{e,s}^{\mathrm{val}}
    =
    \frac{1}{
        N_{\mathrm{val}}
    }
    \sum_{i=1}^{N_{\mathrm{val}}}
    \left|
        \operatorname{Attr}_{i,e,s}
    \right|
    \label{eq:validation_edge_importance}
\end{equation}
denote its normalized mean absolute attribution. Candidate edges are selected
from the top $1\%$ according to their mean importance across discovery seeds.

For scan $i$, edge $e$, and seed $s$, the attribution-derived local sensitivity
is
\begin{equation}
    \ell_{i,e,s}
    =
    \frac{
        \operatorname{Attr}_{i,e,s}
    }{
        x_{i,e}
        -
        x_{e}^{\mathrm{base}}
    },
    \label{eq:local_sensitivity}
\end{equation}
where $x_{e}^{\mathrm{base}}$ is the attribution baseline for edge $e$.

For each discovery seed, the edge direction is obtained from the median
scan-level sensitivity. An edge becomes a frozen explanation claim only when
at least four of the five discovery seeds agree on its sign:
\begin{equation}
    c_e
    =
    (e,d_e),
    \qquad
    d_e\in\{-1,+1\}.
    \label{eq:frozen_edge_claim}
\end{equation}
The frozen claim set is denoted by $\mathcal{C}^{\mathrm{exp}}$. Because claims
are generated separately for each model, different models may produce
different numbers of claims.

\subsubsection{Held-out Confirmation}

For held-out test seed $s$, let
$\operatorname{Top}_{5\%}(a_{\cdot,s}^{\mathrm{test}})$ denote the top $5\%$
of FC edges ranked by test-set mean absolute attribution. Let
$\ell_{e,s}^{\mathrm{test}}$ denote the median test-set local sensitivity of
edge $e$.

Claim $c_e=(e,d_e)$ is confirmed when the edge remains highly attributed and
its test direction agrees with the frozen discovery direction:
\begin{equation}
    I_{e,s}^{\mathrm{conf}}
    =
    \mathbf{1}
    \left[
        e
        \in
        \operatorname{Top}_{5\%}
        \left(
            a_{\cdot,s}^{\mathrm{test}}
        \right)
        \land
        \operatorname{sign}
        \left(
            \ell_{e,s}^{\mathrm{test}}
        \right)
        =
        d_e
    \right].
    \label{eq:confirmation_indicator}
\end{equation}

The seed-level confirmation score is
\begin{equation}
    \operatorname{Confirmation}_{s}
    =
    \frac{1}{
        |\mathcal{C}^{\mathrm{exp}}|
    }
    \sum_{
        e\in\mathcal{C}^{\mathrm{exp}}
    }
    I_{e,s}^{\mathrm{conf}}.
    \label{eq:metric_confirmation}
\end{equation}
Confirmation therefore requires both repeated importance and directional
agreement. It measures explanation reproducibility, but does not by itself
establish that the claimed edge controls the model output.

\subsubsection{Perturbation Faithfulness}

Faithfulness directly perturbs each claimed FC edge. For test scan $i$ and
claim edge $e$, define
\begin{equation}
\begin{aligned}
    x_{i,e}^{+}
    &=
    \operatorname{clip}
    \left(
        x_{i,e}+0.02,
        -0.99,
        0.99
    \right),
    \\
    x_{i,e}^{-}
    &=
    \operatorname{clip}
    \left(
        x_{i,e}-0.02,
        -0.99,
        0.99
    \right).
\end{aligned}
\label{eq:faithfulness_perturbations}
\end{equation}
All non-target FC edges remain unchanged. The central finite-difference
response is
\begin{equation}
    g_{i,e,s}
    =
    \frac{
        f_s(\mathbf{x}_{i,e}^{+})
        -
        f_s(\mathbf{x}_{i,e}^{-})
    }{
        x_{i,e}^{+}
        -
        x_{i,e}^{-}
    }.
    \label{eq:faithfulness_finite_difference}
\end{equation}

The scan-level directional agreement for claim $e$ is
\begin{equation}
    A_{e,s}
    =
    \frac{1}{N_{\mathrm{test}}}
    \sum_{i=1}^{N_{\mathrm{test}}}
    \mathbf{1}
    \left[
        \operatorname{sign}
        \left(
            g_{i,e,s}
        \right)
        =
        d_e
    \right].
    \label{eq:claim_direction_agreement}
\end{equation}

The seed-level faithfulness score is
\begin{equation}
    \operatorname{Faithfulness}_{s}
    =
    \frac{1}{
        |\mathcal{C}^{\mathrm{exp}}|
    }
    \sum_{
        e\in\mathcal{C}^{\mathrm{exp}}
    }
    A_{e,s}.
    \label{eq:metric_faithfulness}
\end{equation}
Thus, Confirmation asks whether the explanation procedure identifies the same
edge and direction again, whereas Faithfulness asks whether intervening on
that edge actually changes the model output in the claimed direction.

\subsection{Domain-alignment Metrics}
\label{app:domain_alignment_metrics}

\subsubsection{Claim Support Rate}

For edge $e$ and seed $s$, define its mean absolute attribution
\begin{equation}
    a_{e,s}
    =
    \frac{1}{N}
    \sum_{i=1}^{N}
    \left|
        \operatorname{Attr}_{i,e,s}
    \right|.
    \label{eq:csr_edge_importance}
\end{equation}
Let
\begin{equation}
    H_s
    =
    \operatorname{Top}_{30\%}
    \left(
        a_{\cdot,s}
    \right)
    \label{eq:top30_edge_set}
\end{equation}
be the top-$30\%$ edge set for seed $s$.

For the prespecified five-seed set $\mathcal{S}_5$ used consistently across
methods, the stable domain claim set is
\begin{equation}
    \mathcal{C}^{\mathrm{dom}}
    =
    \left\{
        e
        \,\middle|\,
        \sum_{s\in\mathcal{S}_5}
        \mathbf{1}
        \left[
            e\in H_s
        \right]
        \geq
        4
    \right\}.
    \label{eq:stable_domain_claims}
\end{equation}

After deduplicating the frozen PubMed evidence, define
\begin{equation}
    \operatorname{Supported}(e)
    =
    \mathbf{1}
    \left[
        \text{at least one applicable supporting source exists for }e
    \right].
    \label{eq:supported_edge_indicator}
\end{equation}
Claim support rate is
\begin{equation}
    \operatorname{CSR}
    =
    \frac{
        \sum_{
            e\in\mathcal{C}^{\mathrm{dom}}
        }
        \operatorname{Supported}(e)
    }{
        |\mathcal{C}^{\mathrm{dom}}|
    }.
    \label{eq:metric_csr}
\end{equation}

CSR measures evidence coverage among stable high-attribution edges. It does
not distinguish exact-connection evidence from broader connection-family
evidence and does not account for attribution magnitude or rank. Because CSR
is computed once from the stable claim set jointly constructed across the
prespecified seeds, it is reported as a value together with its numerator and
denominator rather than as a seed-level mean and standard deviation.

\subsubsection{Attribution--Evidence Alignment}

AEA evaluates whether stronger literature evidence is associated with greater
attribution mass and higher attribution rank:
\begin{equation}
    \operatorname{AEA}_{s}
    =
    \frac{
        \operatorname{nEAM}_{s}
        +
        \operatorname{ERA@30}_{s}
    }{2}.
    \label{eq:metric_aea}
\end{equation}

\paragraph{Edge-level evidence score.}
Evidence records are first deduplicated by PubMed identifier. If the same
publication contributes both connection-family and exact-connection evidence,
only the exact-connection record is retained.

The scope weight is
\begin{equation}
    w_{\mathrm{scope}}
    =
    \begin{cases}
        1.0,
        &
        \text{exact-connection evidence},
        \\
        0.5,
        &
        \text{connection-family evidence}.
    \end{cases}
    \label{eq:evidence_scope_weight}
\end{equation}

For the current age-association query, positive, negative, direction-unspecified,
and mixed age-related findings are treated as supporting evidence because the
claim concerns whether the connection is age-related rather than its specific
FC--age direction. Explicit-null findings are treated as opposing evidence.

Let
\begin{equation}
    W_e^{+}
    =
    \sum_{
        \ell\in\mathcal{B}_e^{+}
    }
    w_{\mathrm{scope},\ell},
    \qquad
    W_e^{-}
    =
    \sum_{
        \ell\in\mathcal{B}_e^{-}
    }
    w_{\mathrm{scope},\ell}.
    \label{eq:evidence_weight_sums}
\end{equation}
The edge evidence score is
\begin{equation}
    s_e
    =
    \operatorname{clip}
    \left(
        \frac{
            W_e^{+}
            -
            W_e^{-}
        }{2},
        -1,
        1
    \right).
    \label{eq:edge_evidence_score}
\end{equation}

\paragraph{Normalized evidence-weighted attribution mass.}
For seed $s$, normalize the complete attribution distribution:
\begin{equation}
    p_{e,s}
    =
    \frac{
        a_{e,s}
    }{
        \sum_k a_{k,s}
    }.
    \label{eq:normalized_attribution_mass}
\end{equation}
The raw evidence-weighted attribution mass is
\begin{equation}
    M_s
    =
    \sum_e
    p_{e,s}s_e.
    \label{eq:raw_evidence_attribution_mass}
\end{equation}

Because different models may have different attribution concentration,
$M_s$ is normalized using rearrangement bounds. Let $M_s^{\max}$ be the score
obtained by optimally matching the largest attribution masses with the largest
evidence scores, and let $M_s^{\min}$ be the score obtained by the reverse
matching. Then
\begin{equation}
    \operatorname{nEAM}_{s}
    =
    \frac{
        M_s
        -
        M_s^{\min}
    }{
        M_s^{\max}
        -
        M_s^{\min}
    }.
    \label{eq:normalized_eam}
\end{equation}
Values near one indicate that attribution mass is allocated close to the
evidence-optimal arrangement, whereas values near zero indicate an allocation
close to the least aligned arrangement.

\paragraph{Evidence-ranking alignment at $30\%$.}
Let $E$ denote the number of FC edges and
\begin{equation}
    K
    =
    \left\lceil
        0.30E
    \right\rceil.
    \label{eq:era_cutoff}
\end{equation}
For the current $81$-region FC representation, $E=3240$ and $K=972$.

Order edges by decreasing attribution,
$(e_1,\ldots,e_E)$, and define
\begin{equation}
    s_e^{+}
    =
    \max
    \left(
        s_e,
        0
    \right).
    \label{eq:positive_evidence_score}
\end{equation}
The discounted cumulative evidence gain is
\begin{equation}
    \operatorname{DCG@K}_{s}
    =
    \sum_{r=1}^{K}
    \frac{
        s_{e_r}^{+}
    }{
        \log_2(r+1)
    }.
    \label{eq:evidence_dcg}
\end{equation}
Let $\operatorname{IDCG@K}$ denote the corresponding value under the ideal
evidence ordering. The ranking-alignment score is
\begin{equation}
    \operatorname{ERA@30}_{s}
    =
    \frac{
        \operatorname{DCG@K}_{s}
    }{
        \operatorname{IDCG@K}
    }.
    \label{eq:era30}
\end{equation}

The current frozen evidence set contains no edge that remains classified as
explicit-null after applicability filtering, so the reported ERA@30 values use
the supporting-evidence ranking above. The implementation additionally
supports opposing evidence by computing its NDCG and using
$1-\operatorname{NDCG}_{\mathrm{opp}}$, weighted by the relative supporting
and opposing evidence strength.

nEAM evaluates how much attribution mass is assigned to evidence-supported
edges, while ERA@30 evaluates whether those edges occur sufficiently early in
the attribution ranking. Their average prevents a model from obtaining a high
domain-alignment score solely through diffuse attribution mass or through a
small number of incidentally highly ranked edges.

\subsection{Controlled-revision Metrics}
\label{app:controlled_revision_metrics}

Controlled-revision metrics are defined only for methods with a matched
transition from a parent predictor to a revised predictor:
\begin{equation}
    f_0^{(s)}
    \longrightarrow
    f_{\mathrm{rev}}^{(s)}.
    \label{eq:matched_refinement_transition}
\end{equation}
They are not defined for independently trained predictive baselines.

\subsubsection{Target Progress}

The current evaluation uses nine revision targets:
\begin{table}[t]
    \centering
    \caption{Revision targets used to compute Progress.}
    \label{tab:progress_targets}
    \scriptsize
    \begin{tabular}{p{0.67\columnwidth}p{0.23\columnwidth}}
        \toprule
        Target statistic & Intended movement \\
        \midrule
        Signed-direction fraction for edge 06\_\_15
        & Toward $1$ \\
        Age-distance representation Spearman correlation
        & Toward $1$ \\
        Temporal direction accuracy
        & Toward $1$ \\
        Prediction-slope error
        & Toward $0$ \\
        Interval--latent-distance Spearman correlation
        & Toward $1$ \\
        Age-bin absolute bias
        & Toward $0$ \\
        Within-subject closer fraction
        & Toward $1$ \\
        Relative-importance ratio over eight target edges
        & Increase \\
        Interval delta-error dispersion
        & Toward $0$ \\
        \bottomrule
    \end{tabular}
\end{table}

Let $u_{s,j}^{0}$ denote target statistic $j$ measured from the matched parent
model for seed $s$, and let $u_{s,j}^{\mathrm{rev}}$ denote the corresponding
statistic after refinement.

For a target with ideal value $u_j^{\star}$, relative gap closure is
\begin{equation}
    g_{s,j}
    =
    \frac{
        \left|
            u_{s,j}^{0}
            -
            u_j^{\star}
        \right|
        -
        \left|
            u_{s,j}^{\mathrm{rev}}
            -
            u_j^{\star}
        \right|
    }{
        \max
        \left(
            \left|
                u_{s,j}^{0}
                -
                u_j^{\star}
            \right|,
            \epsilon
        \right)
    }.
    \label{eq:relative_gap_closure}
\end{equation}

For the relative-importance target, which requires only an increase, define
\begin{equation}
    g_{s,j}
    =
    \frac{
        u_{s,j}^{\mathrm{rev}}
        -
        u_{s,j}^{0}
    }{
        \max
        \left(
            \left|
                u_{s,j}^{0}
            \right|,
            \epsilon
        \right)
    }.
    \label{eq:increase_only_progress}
\end{equation}

Target $j$ is counted as improved when $g_{s,j}>0$. The seed-level Progress
score is
\begin{equation}
    \operatorname{Progress}_{s}
    =
    \frac{1}{9}
    \sum_{j=1}^{9}
    \mathbf{1}
    \left[
        g_{s,j}>0
    \right].
    \label{eq:metric_progress}
\end{equation}
Progress therefore measures the probability that a randomly selected
prespecified target moves in the intended direction, rather than averaging
heterogeneous change magnitudes.

Random-target, opposite-direction, direct-evidence, one-shot, all-eligible, and
complete-loop variants are all evaluated against the same nine targets defined
by the complete \method{} procedure.

\subsubsection{Off-target Drift}

Drift measures whether refinement changes model properties not directly
targeted by the revision constraints. The seven evaluated non-target
dimensions are:
\begin{table}[t]
    \centering
    \caption{Non-target dimensions used to compute Drift.}
    \label{tab:drift_dimensions}
    \scriptsize
    \begin{tabular}{p{0.62\columnwidth}p{0.28\columnwidth}}
        \toprule
        Non-target dimension & Distance \\
        \midrule
        Longitudinal delta MAE
        & Absolute difference \\
        Residual retest correlation
        & Absolute difference \\
        Female--male MAE gap
        & Absolute difference \\
        Individual prediction ranking
        & $(1-\rho_{\mathrm{S}})/2$ \\
        Non-target attribution ranking
        & $(1-\rho_{\mathrm{S}})/2$ \\
        Non-target top-$10\%$ edge set
        & $1-\operatorname{Jaccard}$ \\
        Non-target attribution entropy
        & Absolute difference \\
        \bottomrule
    \end{tabular}
\end{table}
Here, $\rho_{\mathrm{S}}$ denotes Spearman correlation. All explicitly targeted
edges are excluded from attribution-based drift calculations so that intended
changes are not counted as off-target drift.

For dimension $d$, the natural retraining threshold is estimated from
pairwise distances between independently trained prediction-only models:
\begin{equation}
    \tau_d
    =
    Q_{0.95}
    \left(
        \left\{
            D_d
            \left(
                f_0^{(s)},
                f_0^{(s')}
            \right)
            :
            s<s'
        \right\}
    \right).
    \label{eq:drift_null_threshold}
\end{equation}
This threshold estimates the amount of variation that may arise from ordinary
random initialization and retraining without refinement.

For matched seed $s$, define
\begin{equation}
    I_{s,d}^{\mathrm{drift}}
    =
    \mathbf{1}
    \left[
        D_d
        \left(
            f_0^{(s)},
            f_{\mathrm{rev}}^{(s)}
        \right)
        >
        \tau_d
    \right].
    \label{eq:drift_indicator}
\end{equation}
The seed-level Drift score is
\begin{equation}
    \operatorname{Drift}_{s}
    =
    \frac{1}{7}
    \sum_{d=1}^{7}
    I_{s,d}^{\mathrm{drift}}.
    \label{eq:metric_drift}
\end{equation}
Lower values are better. Drift equal to zero means that none of the seven
non-target dimensions changes beyond ordinary prediction-only variability.
Prediction-only models have no matched transition and therefore receive no
Progress or Drift value. Sham continuation uses the same stored artifact as
its parent and consequently has Progress and Drift equal to zero.

\section{Details of Explanation, Verification, and Revision Validity}
\label{app:component_validity}

\subsection{RQ2: Ablation of Reliable Explanation Discovery}
\label{app:rq2_explanation_ablation}

\subsubsection{Evaluation Protocol}

RQ2 evaluates the explanation-discovery component using the
validation-selected Loop-6 predictor. Explanation claims are constructed
exclusively from the validation split and are frozen before any test-set
evaluation.

The discovery and confirmation seed sets are
\begin{equation}
    \mathcal{S}_{\mathrm{disc}}
    =
    \{0,1,2,3,4\},
    \qquad
    \mathcal{S}_{\mathrm{conf}}
    =
    \{5,6,7,8,9\}.
    \label{eq:rq2_seed_sets}
\end{equation}
The discovery stage does not access test subjects. All frozen claims are
evaluated unchanged on the test split and the five independently trained
confirmation models.

For edge $e$, discovery seed $s$, and validation scan $i$, let
$\operatorname{Attr}_{i,e,s}$ denote the edge attribution. Its mean absolute
importance is
\begin{equation}
    a_{e,s}
    =
    \frac{1}{N_{\mathrm{val}}}
    \sum_{i=1}^{N_{\mathrm{val}}}
    \left|
        \operatorname{Attr}_{i,e,s}
    \right|.
    \label{eq:rq2_edge_importance}
\end{equation}

The attribution-derived local sensitivity is
\begin{equation}
    \ell_{i,e,s}
    =
    \frac{
        \operatorname{Attr}_{i,e,s}
    }{
        x_{i,e}
        -
        x_{e}^{\mathrm{base}}
    },
    \label{eq:rq2_local_sensitivity}
\end{equation}
where $x_{e}^{\mathrm{base}}$ is the attribution baseline. The seed-level
direction is obtained from the sign of the median scan-level sensitivity.

\subsubsection{Ablation Protocols}

\paragraph{Single seed, single explainer.}
This protocol uses validation seed $0$ only. Edges are ranked by
$a_{e,0}$, and the top $1\%$ are retained. Each claim is assigned the sign of
its median local sensitivity:
\begin{equation}
    c_e
    =
    (e,d_e),
    \qquad
    d_e
    =
    \operatorname{sign}
    \left(
        \operatorname{median}_{i}
        \ell_{i,e,0}
    \right).
    \label{eq:rq2_single_seed_claim}
\end{equation}
No cross-seed agreement or intervention-based faithfulness requirement is
applied. This protocol produces 33 frozen claims.

\paragraph{Multi-seed without faithfulness gate.}
Normalized edge importance is averaged across the five discovery seeds, and
the top $1\%$ of edges are retained. A claim is accepted only when at least
four of five discovery seeds agree on its local-sensitivity direction:
\begin{equation}
    \sum_{
        s\in\mathcal{S}_{\mathrm{disc}}
    }
    \mathbf{1}
    \left[
        \operatorname{sign}
        \left(
            \operatorname{median}_{i}
            \ell_{i,e,s}
        \right)
        =
        d_e
    \right]
    \geq
    4.
    \label{eq:rq2_multiseed_direction}
\end{equation}
This protocol produces 32 frozen claims.

\paragraph{Full reliable discovery.}
The full protocol begins from the multi-seed claim set and additionally
requires intervention-based faithfulness. For each validation scan and claim
edge, define
\begin{equation}
\begin{aligned}
    x_{i,e}^{+}
    &=
    \operatorname{clip}
    \left(
        x_{i,e}+0.02,
        -0.99,
        0.99
    \right),
    \\
    x_{i,e}^{-}
    &=
    \operatorname{clip}
    \left(
        x_{i,e}-0.02,
        -0.99,
        0.99
    \right).
\end{aligned}
\label{eq:rq2_discovery_perturbation}
\end{equation}
All other FC edges remain unchanged. The finite-difference response is
\begin{equation}
    g_{i,e,s}
    =
    \frac{
        f_s(\mathbf{x}_{i,e}^{+})
        -
        f_s(\mathbf{x}_{i,e}^{-})
    }{
        x_{i,e}^{+}
        -
        x_{i,e}^{-}
    }.
    \label{eq:rq2_discovery_response}
\end{equation}

For edge $e$ and seed $s$, scan-level direction agreement is
\begin{equation}
    A_{e,s}
    =
    \frac{1}{N_{\mathrm{val}}}
    \sum_{i=1}^{N_{\mathrm{val}}}
    \mathbf{1}
    \left[
        \operatorname{sign}
        \left(
            g_{i,e,s}
        \right)
        =
        d_e
    \right].
    \label{eq:rq2_discovery_agreement}
\end{equation}

The claim passes the faithfulness gate when the intervention is evaluable and
$A_{e,s}>0.5$ in at least four of the five discovery seeds:
\begin{equation}
    \sum_{
        s\in\mathcal{S}_{\mathrm{disc}}
    }
    \mathbf{1}
    \left[
        A_{e,s}>0.5
    \right]
    \geq
    4.
    \label{eq:rq2_faithfulness_gate}
\end{equation}
Twenty-nine of the 32 multi-seed claims pass this gate.

\subsubsection{Held-out Confirmation}

For confirmation seed $s$, test-set edge importance is recomputed using
Equation~\eqref{eq:rq2_edge_importance}. A frozen claim is confirmed only when
the same edge remains in the top $5\%$ of test attribution importance and its
test local-sensitivity direction matches the discovery direction:
\begin{equation}
    I_{e,s}^{\mathrm{conf}}
    =
    \mathbf{1}
    \left[
        e
        \in
        \operatorname{Top}_{5\%}
        \left(
            a_{\cdot,s}^{\mathrm{test}}
        \right)
        \land
        \operatorname{sign}
        \left(
            \operatorname{median}_{i}
            \ell_{i,e,s}^{\mathrm{test}}
        \right)
        =
        d_e
    \right].
    \label{eq:rq2_confirmation_indicator}
\end{equation}

For frozen claim set $\mathcal{C}$, seed-level confirmation is
\begin{equation}
    \operatorname{Confirmation}_{s}
    =
    \frac{1}{|\mathcal{C}|}
    \sum_{e\in\mathcal{C}}
    I_{e,s}^{\mathrm{conf}}.
    \label{eq:rq2_confirmation}
\end{equation}

\subsubsection{Held-out Faithfulness}

The same central finite-difference intervention is applied to test scans.
For claim $e$ and confirmation seed $s$, held-out faithfulness is
\begin{equation}
    F_{e,s}
    =
    \frac{1}{N_{\mathrm{test}}}
    \sum_{i=1}^{N_{\mathrm{test}}}
    \mathbf{1}
    \left[
        \operatorname{sign}
        \left(
            g_{i,e,s}^{\mathrm{test}}
        \right)
        =
        d_e
    \right].
    \label{eq:rq2_claim_faithfulness}
\end{equation}
The seed-level score is
\begin{equation}
    \operatorname{Faithfulness}_{s}
    =
    \frac{1}{|\mathcal{C}|}
    \sum_{e\in\mathcal{C}}
    F_{e,s}.
    \label{eq:rq2_faithfulness}
\end{equation}

Both metrics are reported as the mean and sample standard deviation across
$\mathcal{S}_{\mathrm{conf}}$. Claims are never modified after observing
confirmation-seed or test-set results.

\subsection{RQ3: Single-reviewer Audit of Evidence Grounding and Revision Validity}
\label{app:rq3_human_audit}

\subsubsection{Diagnostic-set Construction}

The audit uses 37 evidence-grounding records generated from the Loop-0
baseline. The records form a purposive stratified diagnostic set rather than a
simple random sample.

\begin{table}[t]
\centering
\scriptsize
\caption{Composition of the RQ3 diagnostic audit set.}
\label{tab:rq3_audit_composition}
\begin{tabular}{lc}
\toprule
System sampling stratum & Records \\
\midrule
Insufficient evidence & 12 \\
Scope mismatch or broad relevance & 12 \\
Mixed evidence & 7 \\
Support & 3 \\
Contradiction with model claim & 3 \\
\midrule
Total & 37 \\
\bottomrule
\end{tabular}
\end{table}

For strata containing more than 12 records, the records with the highest
system-assigned domain-evidence strength are selected. Smaller strata include
all available records. The resulting packets contain 572 candidate literature
records, each represented by title, publication year, PubMed identifier, and
extracted evidence sentence.

Because the set is purposively stratified, aggregate audit percentages
describe this diagnostic set and are not interpreted as estimates of
corpus-wide or future-query accuracy.

\subsubsection{Sequential Review Protocol}

The review is conducted in two sequential stages.

\paragraph{Stage 1: Evidence review.}
The reviewer initially sees only the neutral domain question and the candidate
literature records. The reviewer does not see:

\begin{itemize}
    \item the model-side claim;
    \item the automated evidence label;
    \item the sampling stratum;
    \item the proposed action or revision;
    \item downstream model outcomes;
    \item the private sampling key.
\end{itemize}

The reviewer annotates evidence status, population scope, measurement scope,
connection scope, evidence stance, whether the system should abstain from a
precise conclusion, and free-text review notes. The evidence review must be
saved before the revision-review stage is shown.

\paragraph{Stage 2: Revision review.}
After completing the evidence review, the reviewer sees the model claim,
neutral question, candidate domain claim, proposed action, and proposed
revision. The reviewer then judges:

\begin{itemize}
    \item whether the revision is admissible;
    \item whether it preserves the correct model variable;
    \item whether it preserves the correct measurement operator;
    \item whether it preserves the evidence-supported scope;
    \item whether it is the minimum adequate revision.
\end{itemize}

For an unacceptable proposal, the reviewer provides a corrected minimal
revision. Only one reviewer participated; therefore, the analysis reports
system--reviewer agreement and reviewer acceptance rates rather than
inter-reviewer reliability.

\subsubsection{Evidence-status Normalization}

The system and reviewer use related but non-identical evidence taxonomies.
Their labels are compared using the following documented normalization:

\begin{itemize}
    \item system \emph{complete} maps to reviewer \emph{support};
    \item \emph{conflicting exact evidence} maps to \emph{mixed};
    \item \emph{no domain update} maps to \emph{insufficient};
    \item \emph{broad relevance support} agrees when the reviewer accepts
    support only at the corresponding broader level or judges that the exact
    claim should be withheld.
\end{itemize}

The resulting metric is called \emph{evidence-status agreement}, rather than
unqualified evidence accuracy, because the mapping is a documented
post-review normalization.

Let $\widetilde{s}_j^{\mathrm{sys}}$ denote the normalized system label and
$s_j^{\mathrm{rev}}$ the reviewer label. Evidence-status agreement is
\begin{equation}
    \operatorname{Agree}_{\mathrm{status}}
    =
    \frac{1}{37}
    \sum_{j=1}^{37}
    \mathbf{1}
    \left[
        \widetilde{s}_j^{\mathrm{sys}}
        =
        s_j^{\mathrm{rev}}
    \right].
    \label{eq:rq3_status_agreement}
\end{equation}

\subsubsection{Scope and Abstention Agreement}

Strict scope agreement is computed only for records whose system and reviewer
scope categories are directly comparable. The following correspondences are
used:

\begin{itemize}
    \item system \emph{direct}: reviewer population and measurement scopes are
    both exact;
    \item system \emph{exact connection}: reviewer scope is the same exact
    connection;
    \item system \emph{connection family}: reviewer scope is the same
    connection family;
    \item system \emph{none}: reviewer labels the evidence insufficient and
    recommends abstention.
\end{itemize}

Four system records labeled \emph{mixed} have no directly corresponding
reviewer scope category and are excluded, yielding 33 comparable records:
\begin{equation}
    \operatorname{Agree}_{\mathrm{scope}}
    =
    \frac{1}{33}
    \sum_{j\in\mathcal{J}_{\mathrm{scope}}}
    \mathbf{1}
    \left[
        \omega_j^{\mathrm{sys}}
        =
        \omega_j^{\mathrm{rev}}
    \right].
    \label{eq:rq3_scope_agreement}
\end{equation}

For precise-abstention agreement, system states \emph{no domain update} and
\emph{broad relevance support} are treated as withholding a precise
directional or optimization conclusion. Let $b_j^{\mathrm{sys}}$ and
$b_j^{\mathrm{rev}}$ denote the corresponding binary abstention decisions:
\begin{equation}
    \operatorname{Agree}_{\mathrm{abs}}
    =
    \frac{1}{37}
    \sum_{j=1}^{37}
    \mathbf{1}
    \left[
        b_j^{\mathrm{sys}}
        =
        b_j^{\mathrm{rev}}
    \right].
    \label{eq:rq3_abstention_agreement}
\end{equation}

\subsubsection{Revision-validity Metrics}

For each property
\begin{equation}
    p
    \in
    \left\{
        \mathrm{admissible},
        \mathrm{variable},
        \mathrm{operator},
        \mathrm{scope},
        \mathrm{minimal}
    \right\},
\end{equation}
let $I_{j,p}^{\mathrm{rev}}\in\{0,1\}$ denote the reviewer judgment. The
corresponding acceptance rate is
\begin{equation}
    R_p
    =
    \frac{1}{37}
    \sum_{j=1}^{37}
    I_{j,p}^{\mathrm{rev}}.
    \label{eq:rq3_revision_metric}
\end{equation}

The raw reviewer evidence labels are nine support, two mixed, 26 insufficient,
and zero contradiction records; the reviewer recommends abstention for 28
records. The absence of reviewer-labeled contradiction does not conflict with
the three model-contradiction sampling records: those strata indicate that the
literature-supported relation conflicts with the model claim, whereas the
reviewer evaluates whether the literature supports the neutral scientific
question.

\subsubsection{Scope-error Analysis}

Strict scope agreement by system scope is:

\begin{table}[t]
\centering
\scriptsize
\caption{Strict scope agreement by system-assigned evidence scope.}
\label{tab:rq3_scope_breakdown}
\begin{tabular}{lcc}
\toprule
System scope & Agreement & Rate \\
\midrule
None & 12/12 & 100.0\% \\
Exact connection & 1/1 & 100.0\% \\
Direct & 6/8 & 75.0\% \\
Connection family & 4/12 & 33.3\% \\
Mixed & -- & Not comparable \\
\bottomrule
\end{tabular}
\end{table}

The main scope failure occurs for connection-family evidence. In several
records, the retrieved literature establishes only network-level or broad
anatomical relevance and does not justify a connection-family conclusion.

\subsubsection{Rejected-revision Analysis}

Five proposed revisions are judged inadmissible, and all five are
strengthening proposals. Their common failure is that the optimization target
is more specific than the supplied literature evidence:

\begin{itemize}
    \item general test--retest evidence is used to justify age-detrended
    residual consistency;
    \item cohort-average longitudinal tracking is used to require precise
    individual change magnitude;
    \item group-average aging direction is used to require every individual
    pair to follow the same direction;
    \item heterogeneous longitudinal evidence is used to require monotonic
    latent displacement with interval;
    \item prediction repeatability is used to require cross-seed consistency
    of internal representation geometry.
\end{itemize}

These errors do not generally change the diagnosed variable: variable
preservation is accepted for all 37 records. Instead, they reveal
over-compilation, in which related but indirect or population-level evidence
is translated into an overly specific model constraint.

\subsubsection{Interpretation Limitations}

The audit has four limitations. First, it uses one reviewer, so inter-reviewer
agreement and disagreement adjudication cannot be measured. Second, the
records form a purposively stratified diagnostic set rather than a random
sample. Third, evidence-state normalization is documented after review because
the system and reviewer taxonomies are not identical. Fourth, the grounding
records are generated from Loop 0, whereas the RQ2 explanation ablation uses
the validation-selected Loop-6 model. The audit therefore evaluates component
behavior rather than the final model checkpoint specifically.

\section{Details of Targeted Effect and Update Specificity}
\label{app:rq4_targeted_effect}

\subsection{Sequential Refinement Protocol}
\label{app:rq4_protocol}

RQ4 evaluates each refinement strategy over six consecutive scientific-model
updates:
\begin{equation}
    f_{0}^{(m)}
    \longrightarrow
    f_{1}^{(m)}
    \longrightarrow
    \cdots
    \longrightarrow
    f_{6}^{(m)},
    \label{eq:rq4_revision_sequence}
\end{equation}
where $m$ indexes the refinement strategy. At transition $t$, the child model
$f_{t}^{(m)}$ is compared with the corresponding parent
$f_{t-1}^{(m)}$ from the same strategy.

All strategies follow the same sequence of scientific targets and use the same
model architecture, data partitions, optimization budget, and evaluation
operators. The comparison therefore isolates how each strategy constructs,
retains, and realizes explanation revisions.

The scientific target set grows across the refinement trajectory. Early rounds
target signed FC response and age-related latent geometry. Later rounds add
longitudinal direction, prediction calibration, interval-aware representation,
age-specific prediction bias, longitudinal robustness, within-subject
representation consistency, and relative importance of selected FC edges.

\subsection{Compared Refinement Strategies}
\label{app:rq4_strategies}

\paragraph{Sham continuation.}
No explanation revision is applied. The strategy reproduces the
prediction-only reference and establishes the zero-change baseline.

\paragraph{Random-target constraints.}
The same constraint families and revision schedule are retained, but the
scientific targets are replaced by reproducible random targets. This control
tests whether any structured regularization is sufficient to produce apparent
progress.

\paragraph{Opposite-direction constraints.}
The same variables and loss structures are used, but the intended scientific
directions are reversed. This control tests whether successful refinement
depends on the semantic direction of the verified knowledge.

\paragraph{Direct evidence-to-loss.}
Verified evidence is translated directly into generic training losses without
typed minimum projection or response-probe selection. Corresponding constraints
are accumulated across refinement rounds.

\paragraph{One-shot typed revision.}
Each round uses only its newly introduced typed revisions. Previously accepted
constraints are not maintained in later rounds. This strategy tests whether
maintenance of earlier scientific corrections is necessary to prevent
forgetting.

\paragraph{All eligible revisions.}
Every executable revision available at a given round is optimized jointly,
without response probing or sparse portfolio selection. This strategy tests
whether applying more knowledge constraints is preferable to selecting a
compatible subset.

\paragraph{\method{}.}
The complete loop uses typed knowledge-set compilation, minimum-change
projection, response probing, validation-gated portfolio selection, and
maintenance constraints for previously accepted relations.

\subsection{Target Diagnostics}
\label{app:rq4_targets}

The cumulative target set contains nine explanatory diagnostics:

\begin{itemize}
    \item signed FC-response fraction, with ideal value $1$;
    \item age-distance representation correlation, with ideal value $1$;
    \item longitudinal temporal-direction accuracy, with ideal value $1$;
    \item prediction-slope error, with ideal value $0$;
    \item interval--latent-distance correlation, with ideal value $1$;
    \item age-bin absolute prediction bias, with ideal value $0$;
    \item interval delta-error dispersion, with ideal value $0$;
    \item within-subject closer fraction, with ideal value $1$;
    \item relative-importance ratio for selected FC edges, which should
    increase.
\end{itemize}

Let $\mathcal{T}_t$ denote the target diagnostics active at transition $t$, and
let $\mathcal{U}_t$ denote the diagnostics not directly targeted by the current
portfolio.

\subsection{Normalized Target-gap Closure}
\label{app:rq4_ntgc}

For target $k$ with ideal value $r_k$, define the parent and child gaps
\begin{equation}
    d_{s,t,k}^{P}
    =
    \left|
        z_{s,t-1,k}
        -
        r_k
    \right|,
    \qquad
    d_{s,t,k}^{C}
    =
    \left|
        z_{s,t,k}
        -
        r_k
    \right|.
    \label{eq:rq4_target_gaps}
\end{equation}

The normalized closure is
\begin{equation}
    q_{s,t,k}
    =
    \frac{
        d_{s,t,k}^{P}
        -
        d_{s,t,k}^{C}
    }{
        d_{s,t,k}^{P}
        +
        \epsilon
    }.
    \label{eq:rq4_target_closure}
\end{equation}

For a target requiring only an increase, such as relative importance, we use
\begin{equation}
    q_{s,t,k}
    =
    \frac{
        z_{s,t,k}
        -
        z_{s,t-1,k}
    }{
        \left|
            z_{s,t-1,k}
        \right|
        +
        \epsilon
    }.
    \label{eq:rq4_increase_closure}
\end{equation}

To prevent extreme values when the parent is already close to the target,
closure is clipped:
\begin{equation}
    \widetilde q_{s,t,k}
    =
    \operatorname{clip}
    \left(
        q_{s,t,k},
        -1,
        1
    \right).
    \label{eq:rq4_clipped_closure}
\end{equation}

Normalized target-gap closure is
\begin{equation}
    \operatorname{NTGC}_{s,t}
    =
    \frac{1}{
        |\mathcal{T}_t|
    }
    \sum_{k\in\mathcal{T}_t}
    \widetilde q_{s,t,k}.
    \label{eq:rq4_ntgc}
\end{equation}
Positive NTGC indicates net movement toward the active scientific targets,
while negative NTGC indicates net movement away from them.

\subsection{Target Improvement Probability}
\label{app:rq4_tip}

Target improvement probability records only whether each target moves in the
correct direction:
\begin{equation}
    \operatorname{TIP}_{s,t}
    =
    \frac{1}{
        |\mathcal{T}_t|
    }
    \sum_{k\in\mathcal{T}_t}
    \mathbf{1}
    \left[
        q_{s,t,k}>0
    \right].
    \label{eq:rq4_tip}
\end{equation}
Unlike NTGC, TIP does not account for the magnitude of improvement or
deterioration.

\subsection{Normalized Off-target Change}
\label{app:rq4_notc}

For non-target diagnostic $u\in\mathcal{U}_t$, the normalized parent--child
change is denoted by $\delta_{s,t,u}$. For bounded proportions, probabilities,
and slope errors, we use absolute difference:
\begin{equation}
    \delta_{s,t,u}
    =
    \left|
        z_{s,t,u}
        -
        z_{s,t-1,u}
    \right|.
    \label{eq:rq4_absolute_change}
\end{equation}

For correlations with range $[-1,1]$, the difference is normalized by the
range:
\begin{equation}
    \delta_{s,t,u}
    =
    \frac{
        \left|
            z_{s,t,u}
            -
            z_{s,t-1,u}
        \right|
    }{2}.
    \label{eq:rq4_correlation_change}
\end{equation}

For positive-scale quantities such as errors or ratios, we use
\begin{equation}
    \delta_{s,t,u}
    =
    \frac{
        \left|
            z_{s,t,u}
            -
            z_{s,t-1,u}
        \right|
    }{
        \max
        \left(
            |z_{s,t-1,u}|,
            10^{-3}
        \right)
    }.
    \label{eq:rq4_relative_change}
\end{equation}

Normalized off-target change is
\begin{equation}
    \operatorname{NOTC}_{s,t}
    =
    \frac{1}{
        |\mathcal{U}_t|
    }
    \sum_{u\in\mathcal{U}_t}
    \delta_{s,t,u}.
    \label{eq:rq4_notc}
\end{equation}
NOTC measures unintended change magnitude and does not assign a positive value
to accidental improvement of a non-target diagnostic.

\subsection{Material Off-target Change Rate}
\label{app:rq4_mocr}

An off-target diagnostic is considered materially changed when its normalized
change exceeds $0.01$:
\begin{equation}
    \operatorname{MOCR}_{s,t}
    =
    \frac{1}{
        |\mathcal{U}_t|
    }
    \sum_{u\in\mathcal{U}_t}
    \mathbf{1}
    \left[
        \delta_{s,t,u}
        >
        0.01
    \right].
    \label{eq:rq4_mocr}
\end{equation}
NOTC measures the average magnitude of collateral change, whereas MOCR
measures how broadly that change is distributed across non-target diagnostics.

\subsection{Targeted Response Specificity}
\label{app:rq4_trs}

We retain only positive target closure:
\begin{equation}
    q_{s,t}^{+}
    =
    \max
    \left(
        \operatorname{NTGC}_{s,t},
        0
    \right).
    \label{eq:rq4_positive_closure}
\end{equation}

Targeted response specificity is
\begin{equation}
    \operatorname{TRS}_{s,t}
    =
    \frac{
        q_{s,t}^{+}
    }{
        q_{s,t}^{+}
        +
        \operatorname{NOTC}_{s,t}
        +
        \epsilon
    }.
    \label{eq:rq4_trs}
\end{equation}
TRS approaches one when target improvement is large relative to off-target
change. If NTGC is negative, TRS is zero.

\subsection{Predictive Preservation Rate}
\label{app:rq4_ppr}

For each transition, prediction is considered preserved when validation MAE
does not increase:
\begin{equation}
    p_{s,t}
    =
    \mathbf{1}
    \left[
        \operatorname{MAE}^{\mathrm{val}}_{s,t}
        \leq
        \operatorname{MAE}^{\mathrm{val}}_{s,t-1}
    \right].
    \label{eq:rq4_prediction_preservation}
\end{equation}

The seed-level predictive preservation rate is
\begin{equation}
    \operatorname{PPR}_{s}
    =
    \frac{1}{6}
    \sum_{t=1}^{6}
    p_{s,t}.
    \label{eq:rq4_ppr}
\end{equation}
The sham strategy obtains PPR equal to one because it does not change the
model; this is not interpreted as successful scientific refinement.

\subsection{Directional Persistence Rate}
\label{app:rq4_dpr}

Let
\begin{equation}
    \mathcal{A}_{s,t}^{\mathrm{val}}
    =
    \left\{
        k
        \,\middle|\,
        q_{s,t,k}^{\mathrm{val}}>0
    \right\}
    \label{eq:rq4_validation_improved_targets}
\end{equation}
denote targets that improve on validation data.

Directional persistence measures how many of these targets also improve on
test data:
\begin{equation}
    \operatorname{DPR}_{s,t}
    =
    \frac{
        \sum_{
            k\in
            \mathcal{A}_{s,t}^{\mathrm{val}}
        }
        \mathbf{1}
        \left[
            q_{s,t,k}^{\mathrm{test}}>0
        \right]
    }{
        \left|
            \mathcal{A}_{s,t}^{\mathrm{val}}
        \right|
    }.
    \label{eq:rq4_dpr}
\end{equation}
When no target improves on validation data, the transition-level DPR is set to
zero. DPR measures directional persistence, not improvement magnitude.

\subsection{Aggregation}
\label{app:rq4_aggregation}

Each metric is first computed separately for every strategy, model run, and
refinement transition. Target and non-target diagnostics are averaged within a
transition, transitions are then weighted equally within a model run, and the
reported mean and sample standard deviation are computed across model runs.

Equal weighting across transitions prevents later rounds with larger target
sets from dominating earlier rounds. NOTC, MOCR, and TRS are averaged only
over transitions containing at least one non-target diagnostic.

\subsection{Interpretation}
\label{app:rq4_interpretation}

NTGC and TIP characterize target movement, but do not indicate whether the
update is selective. NOTC and MOCR characterize collateral change, but do not
measure whether the intended target improved. TRS combines these two aspects.
PPR evaluates predictive preservation, while DPR evaluates whether
validation-selected target movement persists on test data.

Accordingly, a successful refinement strategy should not be selected from one
metric alone. A method may produce large NTGC while causing widespread
off-target change, or low drift simply because it makes no meaningful update.
The complete evaluation therefore considers target movement, specificity,
prediction preservation, and persistence jointly.

\section{Operational Details of the Component Stress Tests}
\label{app:rq6_ablation}

\section{Operational Details of the Component Ablation}
\label{app:rq6_ablation}

\subsection{End-to-end Ablation Protocol}
\label{app:rq6_scope}

Each component ablation is executed as a complete end-to-end refinement
pipeline. All variants use the same subject-disjoint data partitions,
prespecified random seeds, initial prediction-only checkpoints, maximum
refinement rounds, optimization budgets, validation criteria, and evaluation
operators as the full \method{} configuration. The variants differ only in the
component explicitly identified by the ablation.

For every variant, model explanation discovery, literature retrieval and
verification, revision construction, model optimization, and validation-based
promotion are rerun under the corresponding ablated configuration. The
reported models are therefore independently trained outputs of their
respective pipelines rather than post-hoc modifications of artifacts produced
by the full system.

The same matched random seeds are used across the full method and every
ablation variant. Reported values are the mean and sample standard deviation
across these matched runs. The test set is not used for query construction,
revision selection, response probing, model promotion, or refinement-loop
termination.

\subsection{Full Configuration}
\label{app:rq6_full}

The full configuration retains only model-side relations that satisfy the
reproducibility and faithfulness requirements. Each retained relation is
converted into a model-blind scientific question, and the retrieved evidence
is verified with respect to population, measurement, anatomical, connection,
temporal, and directional scope.

Applicable evidence is compiled into a typed admissible set in the same
explanation space as the originating model relation. The current relation is
minimally projected onto this set, and the resulting revision preserves the
originating model variable and measurement operator. Candidate realization
losses are response-probed before full training, and a compatible subset is
selected for joint optimization. A candidate model is promoted only when it
satisfies the prediction, target-movement, and off-target validation criteria.

\subsection{Ablation Definitions}
\label{app:rq6_variants}

\paragraph{Without reliability filtering.}
This variant removes the reproducibility and faithfulness gate applied before
literature verification. Model-side relations that are otherwise evaluable
may enter the verification and refinement pipeline without satisfying the
full reliability requirements. All subsequent literature-verification,
revision-construction, training, and promotion stages remain unchanged.

\paragraph{Without model-blind questions.}
The full method withholds the model-observed direction and magnitude when
constructing the scientific question. This variant exposes the current
model-side relation to the question-generation and evidence-retrieval stages.
All other verification, projection, realization, and promotion procedures are
kept unchanged.

\paragraph{Without scope-aware verification.}
This variant removes the explicit compatibility assessment between the
model-side scope and the evidence-supported scope. Retrieved evidence is
compiled with respect to the originating model scope without applying the full
population, measurement, anatomical, connection, and temporal-scope
restrictions used by the complete method. The remaining projection, training,
and validation procedures are unchanged.

\paragraph{Without minimum projection.}
The full method selects the nearest relation in the verified admissible set.
This variant retains the same verified evidence and admissible relation type
but replaces the minimum-change projection with a stronger
knowledge-consistent target. The candidate is otherwise trained and evaluated
using the same realization and promotion procedures.

\paragraph{Without variable--operator preservation.}
The full method requires every executable revision to inherit the model
variable and measurement operator of the originating explanation. This
variant removes this binding requirement, allowing a verified relation to be
realized through a compatible training quantity that is not required to
preserve both the original variable and operator. All remaining refinement
stages are unchanged.

\paragraph{Without response probing.}
This variant skips the short candidate-response probe performed before full
optimization. All otherwise eligible typed revisions and their corresponding
surrogate losses proceed directly to full retraining under the same
optimization budgets. Importantly, this variant retains typed revision,
minimum projection, variable--operator preservation, and validation-based
promotion; it differs from the complete method only in the removal of
pretraining response probing.

\paragraph{Without sparse portfolio selection.}
The complete method selects a compatible subset of the revisions that pass
response probing. This variant jointly applies all revisions that satisfy the
reliability, evidence, scope, projection, and response-probe requirements.
The individual revision definitions, optimization budgets, and validation
promotion criteria remain unchanged.

\subsection{Evaluation Protocol}
\label{app:rq6_evaluation}

Every ablation is evaluated using the same prediction, explanation, and
controlled-revision metrics as the complete method whenever the corresponding
quantity is defined. Table~\ref{tab:ablation_results} reports MAE, PCC,
held-out explanation Confirmation, target Progress, and off-target Drift
because these metrics are available for all variants.

Comparisons are paired by random seed. For each seed, the complete method and
all ablation variants begin from the corresponding matched prediction-only
initialization and are evaluated on the same validation and test subjects.
Consequently, differences across rows reflect the removal or modification of
the indicated component under an otherwise matched end-to-end protocol.

\subsection{Ablation Results}
\label{app:rq6_failure_modes}

Removing reliability filtering reduces held-out explanation Confirmation and
target Progress, indicating that reproducibility and faithfulness screening
provide a more reliable basis for downstream model revision. Exposing the
model-side conclusion during question construction also reduces prediction
and target Progress, consistent with the role of model-blind questions in
preventing the verification stage from simply reinforcing the current model
relation.

Removing scope-aware verification retains relatively high Confirmation but
produces substantially lower Progress. This result shows that a reproducible
model explanation is not sufficient for intervention when the retrieved
evidence does not apply at the population, measurement, anatomical, or
temporal resolution required by the proposed revision.

The non-minimal variant produces poorer prediction and explanation
Confirmation than the complete method, showing that a stronger
knowledge-consistent intervention is not necessarily preferable to the
smallest evidence-supported revision. Removing variable--operator preservation
also reduces Progress, supporting the requirement that the executable
constraint act on the same explanatory quantity that generated the verified
hypothesis.

The variants without response probing and without sparse portfolio selection
retain relatively high target Progress but produce greater off-target Drift
than the complete method. These results indicate that response probing and
portfolio selection do not merely increase target movement; they help identify
revisions that are controllable and jointly compatible while limiting
unintended changes in non-target explanatory behavior.

\subsection{Interpretation}
\label{app:rq6_limitations}

Because every variant reruns the complete pipeline under matched random seeds
and experimental settings, the study provides an end-to-end component
ablation of the implemented system. The results support the operational role
of each component within the present dataset, predictor, verification
pipeline, and refinement schedule. As with any ablation study, the numerical
effect of removing a component may depend on the dataset, backbone,
hyperparameter configuration, and available evidence corpus, and should not
be interpreted as a universal effect size outside the evaluated setting.
\end{document}